# Survey and cross-benchmark comparison of remaining time prediction methods in business process monitoring

ILYA VERENICH, Queensland University of Technology, Australia and University of Tartu, Estonia
MARLON DUMAS, University of Tartu, Estonia
MARCELLO LA ROSA, University of Melbourne, Australia
FABRIZIO MARIA MAGGI, University of Tartu, Estonia
IRENE TEINEMAA, University of Tartu, Estonia

Predictive business process monitoring methods exploit historical process execution logs to generate predictions about running instances (called cases) of a business process, such as the prediction of the outcome, next activity or remaining cycle time of a given process case. These insights could be used to support operational managers in taking remedial actions as business processes unfold, e.g. shifting resources from one case onto another to ensure this latter is completed on time. A number of methods to tackle the remaining cycle time prediction problem have been proposed in the literature. However, due to differences in their experimental setup, choice of datasets, evaluation measures and baselines, the relative merits of each method remain unclear. This article presents a systematic literature review and taxonomy of methods for remaining time prediction in the context of business processes, as well as a cross-benchmark comparison of 16 such methods based on 16 real-life datasets originating from different industry domains.

## 1 INTRODUCTION

Business process monitoring is concerned with the analysis of events produced during the execution of a business process in order to assess the fulfilment of its compliance requirements and performance objectives [15]. Monitoring can take place offline (e.g., based on periodically produced reports) or online via dashboards displaying the performance of currently running cases of a process [6].

*Predictive* business process monitoring refers to a family of online process monitoring techniques that seek to predict the future state or properties of ongoing executions of a process based on models extracted from historical process executions, stored in so-called event logs. A wide range of predictive business process monitoring methods have been proposed to predict for example compliance violations [24, 25], the next activity or the remaining sequence of activities of a process instance [16, 53], or quantitative process performance indicators such as the remaining cycle time of a process instance [37, 38, 52]. These predictions can be used to alert process workers to problematic process instances or to support resource allocation decisions, e.g. to allocate additional resources to instances that are at risk of a deadline violation.

In this paper, we focus on the problem of predicting time-related properties of ongoing process instances (called process cases), such as remaining time, completion time and case duration. As a result, the contribution of this paper is three-fold: (i) performing a systematic literature review of predictive process monitoring methods for time-related properties; (ii) providing a taxonomy of

Authors' addresses: Ilya Verenich, Queensland University of Technology, 2 George St, Brisbane, QLD, 4000, Australia, University of Tartu, Tartu, Estonia, ilya.verenich@qut.edu.au; Marlon Dumas, University of Tartu, Tartu, Estonia, marlon.dumas@ut.ee; Marcello La Rosa, University of Melbourne, Melbourne, VIC, Australia, marcello.larosa@unimelb.edu.au; Fabrizio Maria Maggi, University of Tartu, Tartu, Estonia, f.m.maggi@ut.ee; Irene Teinemaa, University of Tartu, Tartu, Estonia, irheta@ut.ee.

existing methods; and (iii) performing a comparative experimental evaluation of some representative methods, using a cross-benchmark of predictive monitoring tasks based on a range of real-life event logs. Our work builds upon a recent survey of predictive business process monitoring techniques [26]. However, our work focuses on techniques that predict remaining process time as output, while [26] provides a general-purpose survey, including predictions of time-related properties, probability of a risk or prediction of the next event, among others. As such, our taxonomy aims to better capture specificities of remaining time prediction methods. Secondly, apart from the survey, we perform a cross-benchmark of 16 representative methods based on 16 real-world event logs, extracted from different industry domains. Ultimately, we seek to provide recommendations to business process practitioners as to which method would be more suitable in any particular scenario.

The rest of the paper is organized as follows. Section 2 summarizes some basic concepts in the area of predictive process monitoring. Section 3 describes the search and selection of relevant studies. Section 4 surveys the selected studies and provides a taxonomy to classify them, while Section 5 justifies the selection of methods to be evaluated in our benchmark. Section 6 reports on the benchmark of the selected studies while Section 7 identifies threats to validity. Finally, Section 8 summarizes the findings and identifies directions for future work.

## 2 BACKGROUND

Predictive process monitoring is a multi-disciplinary area that draws concepts from process mining on one side, and machine learning on the other. In this section, we introduce concepts from the aforementioned disciplines that are used in later sections of this paper.

### 2.1 Process mining

Business processes are generally supported by enterprise systems that record data about each individual execution of a process (also called a *case*). Process mining [50] is a research area within business process management that is concerned with deriving useful insights from process execution data (called event logs). Process mining techniques are able to support various stages of business process management tasks, such as process discovery, analysis, redesign, implementation and monitoring [50]. In this subsection, we introduce the key process mining concepts.

Each case consists of a number of *events* representing the execution of activities in a process. Each event has a range of attributes of which three are mandatory, namely (i) *case identifier* specifying which case generated this event, (ii) the *event class* (or *activity name*) indicating which activity the event refers to and (iii) the *timestamp* indicating when the event occurred[1]. An event may carry additional attributes in its payload. For example, in a patient treatment process in a hospital, the name of a responsible nurse may be recorded as an attribute of an event referring to activity "Perform blood test". These attributes are referred to as *event attributes*, as opposed to *case attributes* that belong to the case and are therefore shared by all events relating to that case. For example, in a patient treatment process, the age and gender of a patient can be treated as a case attribute. In other words, case attributes are static, i.e. their values do not change throughout the lifetime of a case, as opposed to attributes in the event payload, which are dynamic as they change from an event to the other.

Formally, an event record is defined as follows:

*Definition 2.1 (Event).* An *event* is a tuple $(a, c, t, (d_1, v_1), \ldots, (d_m, v_m))$ where $a$ is the activity name, $c$ is the case identifier, $t$ is the timestamp and $(d_1, v_1) \ldots, (d_m, v_m)$ (where $m \geq 0$) are the event or case attributes and their values.

[1] Hereinafter, we refer to the *completion* timestamp unless otherwise noted.

Let $\mathcal{E}$ be the event universe, i.e., the set of all possible event classes, and $\mathcal{T}$ the time domain. Then there is a function $\pi_{\mathcal{T}} \in \mathcal{E} \rightarrow \mathcal{T}$ that assigns timestamps to events.

The sequence of events generated by a given case forms a *trace*. Formally,

*Definition 2.2 (Trace).* A *trace* is a non-empty sequence $\sigma = \langle e_1, \ldots, e_n \rangle$ of events such that $\forall i \in [1..n], e_i \in \mathcal{E}$ and $\forall i, j \in [1..n]$ $e_i.c = e_j.c$. In other words, all events in the trace refer to the same case.

A set of *completed traces* (i.e. traces recording the execution of completed cases) comprises an *event log*.

*Definition 2.3 (Event log).* An event log $L$ is a set of completed traces, i.e., $L = \{\sigma_i : \sigma_i \in \mathcal{S}, 1 \leq i \leq K\}$, where $\mathcal{S}$ is the universe of all possible traces and $K$ is the number of traces in the event log.

As a running example, let us consider an extract of an event log originating from an insurance claims handling process (Table 1). The activity name of the first event in case 1 is *A*, it occurred on *1/1/2017* at *9:13AM*. The additional event attributes show that the cost of the activity was 15 units and the activity was performed by *John*. These two are event attributes. The events in each case also carry two case attributes: the age of the applicant and the channel through which the application has been submitted. The latter attributes have the same value for all events of a case.

Table 1. Extract of an event log

| Case | Case attributes | | Event attributes | | | |
|---|---|---|---|---|---|---|
| ID | Channel | Age | Activity | Time | Resource | Cost |
| 1 | Email | 37 | A | 1/1/2017 9:13:00 | John | 15 |
| 1 | Email | 37 | B | 1/1/2017 9:14:20 | Mark | 25 |
| 1 | Email | 37 | D | 1/1/2017 9:16:00 | Mary | 10 |
| 1 | Email | 37 | F | 1/1/2017 9:18:05 | Kate | 20 |
| 1 | Email | 37 | G | 1/1/2017 9:18:50 | John | 20 |
| 1 | Email | 37 | H | 1/1/2017 9:19:00 | Kate | 15 |
| 2 | Email | 52 | A | 2/1/2017 16:55:00 | John | 25 |
| 2 | Email | 52 | D | 2/1/2017 17:00:00 | Mary | 25 |
| 2 | Email | 52 | B | 3/1/2017 9:00:00 | Mark | 10 |
| 2 | Email | 52 | F | 3/1/2017 9:01:50 | Kate | 15 |

Data attributes are typically divided into numeric (quantitative) and categorical (qualitative) data type [44]. Each data type requires different preprocessing to be used in a predictive model. With respect to the running example, numeric attributes are *Age Cost* and *Time* (relative), while categorical attributes are *Channel, Activity* and *Resource*.

As we aim to make predictions for traces of incomplete cases, rather than for traces of completed cases, we define a function that returns the first $k$ events of a trace of a (completed) case.

*Definition 2.4 (Prefix function).* Given a trace $\sigma = \langle e_1, \ldots, e_n \rangle$ and a positive integer $k \leq n$, $hd^k(\sigma) = \langle e_1, \ldots, e_k \rangle$.

For example, for a sequence $\sigma_1 = \langle a, b, c, d, e \rangle$, $hd^2(\sigma_1) = \langle a, b \rangle$.

The application of a prefix function will result in a *prefix log*, where each possible prefix of an event log becomes a trace.

*Definition 2.5 (Prefix log).* Given an event log $L$, its prefix log $L^*$ is the event log that contains all prefixes of $L$, i.e., $L^* = \{hd^k(\sigma) : \sigma \in L, 1 \leq k \leq |\sigma|\}$.

For example, a complete trace consisting of three events would correspond to three traces in the prefix log – the partial trace after executing the first, the second and the third event.

### 2.2 Machine learning

Machine learning is a research area of computer science concerned with the discovery of models, patterns, and other regularities in data [28]. Closely related to machine learning is data mining. Data mining is the "core stage of the *knowledge discovery process* that is aimed at the extraction of interesting – non-trivial, implicit, previously unknown and potentially useful – information from data in large databases" [17]. Data mining techniques focus more on exploratory data analysis, i.e. discovering unknown properties in the data, and are often used as a preprocessing step in machine learning to improve model accuracy.

A machine learning system is characterized by a learning algorithm and training data. The algorithm defines a process of learning from information extracted, usually as *features vectors*, from the training data. In this work, we will deal with *supervised* learning, meaning that training data is labeled, i.e. represented in the following form:

$$D = \{(\mathbf{x}_1, y_1), \ldots, (\mathbf{x}_n, y_n) : n \in \mathbb{N}\}, \tag{1}$$

where $\mathbf{x}_i \in \mathcal{X}$ are $m$-dimensional feature vectors ($m \in \mathbb{N}$) and $y_i \in \mathcal{Y}$ are the corresponding labels, i.e. values of the target variable.

Feature vectors extracted from the labeled training data are used to fit a predictive model that would assign labels on new data given labeled training data while minimizing error and model complexity. In other words, a model generalizes the pattern identified in the training data, providing a mapping $\mathcal{X} \to \mathcal{Y}$. The labels can be either continuous, e.g. cycle time of an activity, or discrete, e.g. loan grade. In the former case, the model is referred to as regression; while in the latter case we are talking about a classification model.

From a probabilistic perspective, the machine learning objective is to infer a conditional distribution $P(\mathcal{Y}|\mathcal{X})$. A standard approach to tackle this problem is to represent the conditional distribution with a parametric model, and then to obtain the parameters using a training set containing $\{\mathbf{x}_n, y_n\}$ pairs of input feature vectors with corresponding target output vectors. The resulting conditional distribution can be used to make predictions of $y$ for new values of $\mathbf{x}$. This is called a *discriminative* approach, since the conditional distribution discriminates between the different values of $y$ [3].

Another approach is to calculate the joint distribution $P(\mathcal{X}, \mathcal{Y})$, expressed as a parametric model, and then apply it to find the conditional distribution $P(\mathcal{Y}|\mathcal{X})$ to make predictions of $y$ for new values of $\mathbf{x}$. This is commonly known as a generative approach since by sampling from the joint distribution one can generate synthetic examples of the feature vector $\mathbf{x}$ [3].

To sum up, *discriminative* approaches try to define a (hard or soft) decision boundary that divides the feature space into areas containing feature vectors belonging to the same class (see Figure 1). In contrast, *generative* approaches first model the probability distributions for each class and then label a new instance as a member of a class whose model is most likely to have generated the instance [3].

### 2.3 Predictive process monitoring

Given a partial trace of a process case, we want to predict a process performance measure in the future, e.g. time of a case until completion (or remaining time). This task is sketched in Figure 2. A *Prediction point* is the point in time where the prediction takes place. A *Predicted point* is a point in time in the future where the performance measure has the predicted value. A prediction is thus based on the knowledge of the predictor on the history of the process execution to the prediction point and the future to the predicted point. The former is warranted by the predictor's *memory* and the latter is based on the predictor's *forecast* (i.e. predicting the future based on trend and seasonal pattern analysis). Finally, the prediction is performed based on a *reasoning method*.

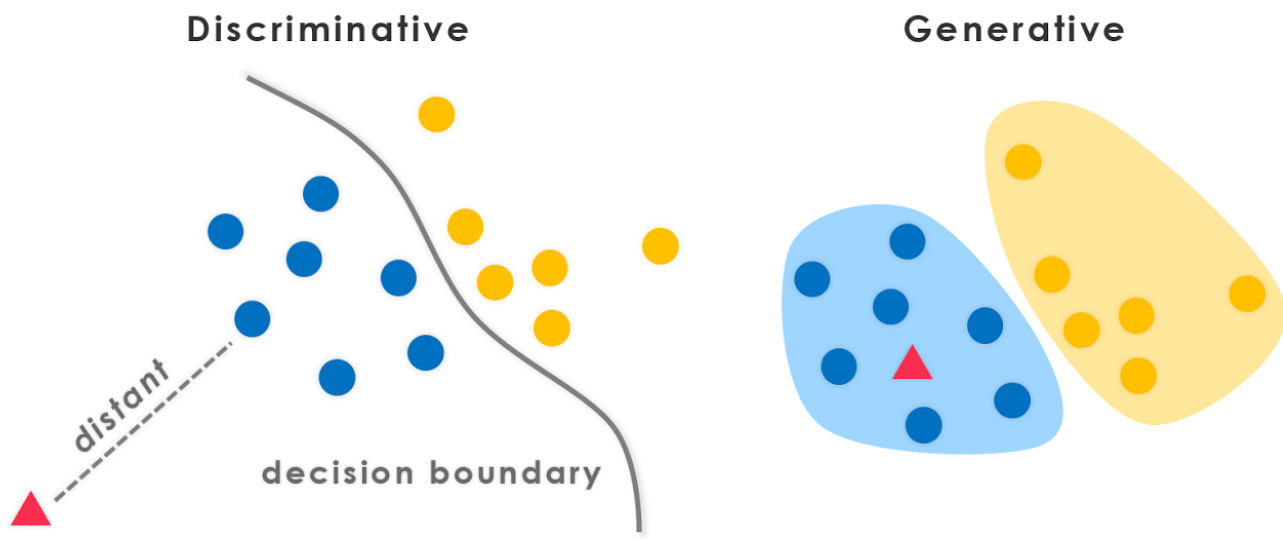

Fig. 1. Discriminative and generative models [30].

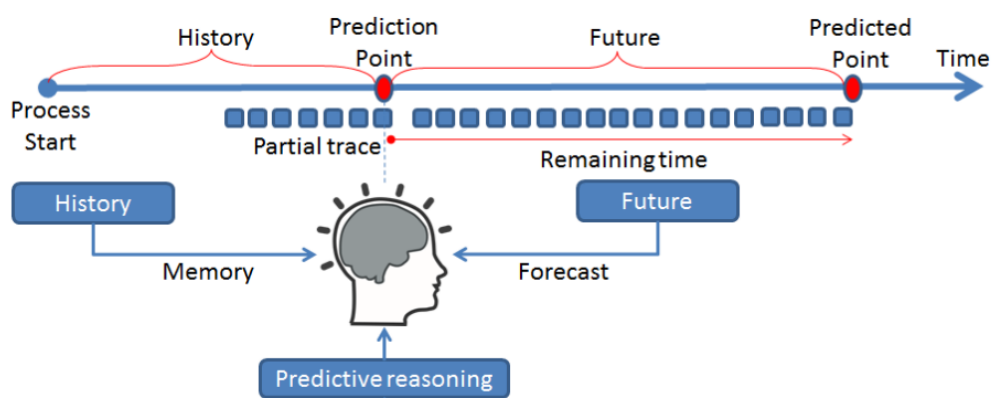

Fig. 2. Overview of predictive process monitoring.

Since in real-life processes the amount of uncertainty increases over time (*cone of uncertainty*), the prediction task becomes more difficult and generally less accurate. As such, predictions are typically made up to a specific point of time in the future, i.e. the time horizon *h*. The choice of *h* depends on how fast the process evolves and on the prediction objectives.

## 3 SEARCH METHODOLOGY

In order to retrieve and select studies for our survey and benchmark, we conducted a *Systematic Literature Review* (SLR) according to the approach described in [23]. We started by specifying the research questions. Next, guided by these goals, we developed relevant search strings for querying a database of academic papers. We applied inclusion and exclusion criteria to the retrieved studies in order to filter out irrelevant ones, and last, we divided all relevant studies into primary and subsumed ones based on their contribution.

### 3.1 Research questions

The purpose of this survey is to define a taxonomy of methods for predictive monitoring of remaining time of business processes. The decision to focus on remaining time is to have a well-delimited and manageable scope, given the richness of the literature in the broader field of predictive process monitoring, and the fact that other predictive process monitoring tasks might rely on different techniques and evaluation measures.

In line with the selected scope, in this benchmark, we aim to answer the following research questions:

RQ1 What methods exist for predictive monitoring of remaining time of business processes?
RQ2 How to classify methods for predictive monitoring of remaining time?

RQ3 What type of data has been used to evaluate these methods, and from which application domains?
RQ4 What tools are available to support these methods?
RQ5 What is the relative performance of these methods?

RQ1 is the core research question, which aims at identifying existing methods to perform predictive monitoring of remaining time. With RQ2, we aim to identify a set of classification criteria on the basis of input data required (e.g. input log) and the underlying predictive algorithms. RQ3 explores what tool support the different methods have, while RQ4 investigates how the methods have been evaluated and in which application domains. Finally, with RQ5, we aim to cross-benchmark existing methods using a set of real-life logs.

### 3.2 Study retrieval

Existing literature in predictive business process monitoring was searched for using Google Scholar, a well-known electronic literature database, as it covers all relevant databases such as ACM Digital Library and IEEE Xplore, and also allows searching within the full text of a paper.

Our search methodology is based on the one proposed in [26], with few variations. Firstly, we collected publications using more specific search phrases, namely "predictive process monitoring", "predictive business process monitoring", "predict (the) remaining time", "remaining time prediction" and "predict (the) remaining * time". The latter is included since some authors refer to the prediction of the remaining *processing* time, while other may call it remaining *execution* time and so on. We retrieved all studies that contained at least one of the above phrases in the title or in the full text of the paper. The search was conducted in March 2018 to include all papers published between 2005 and 2017.

The initial search returned **670** unique results which is about 3 times more than the ones found in [26], owing to the differences in search methodologies (Table 2). Figure 3 shows how the studies are distributed over time. We can see that the interest in the topic of predictive process monitoring grows over time with a sharp increase over the past few years.

Table 2. Comparison of our search methodology with [26]

| | Method in [26] | Our method |
|---|---|---|
| Keywords | 1. "predictive monitoring" AND "business process"<br>2. "business process" AND "prediction" | 1. "predictive process" monitoring<br>2. "predictive business process monitoring"<br>3. "predict (the) remaining time"<br>4. "remaining time prediction"<br>5. "predict (the) remaining * time" |
| Search scope | Title, abstract, keywords | Title, full text |
| Min number of citations | 5 (except 2016-2017 papers) | 5 (except 2017 papers) |
| Years covered | 2010-2016 | 2005-2017 |
| Papers found after filtering | 41 | 53 |
| Snowballing applied | No | Yes, one-hop |

In order to consider only relevant studies, we designed a range of exclusion criteria to assess the relevance of the studies. First, we excluded those papers not related to the process mining field, written in languages other than English or papers with inaccessible full text. Additionally, to contain the reviewing effort, we have only included papers that have been cited at least five times. An exception has been made for papers published in 2017 – as many of them have not had a chance to accumulate the necessary number of citations, we required only one citation for them. Thus, after the first round of filtering, a total of **53** publications were considered for further evaluation.

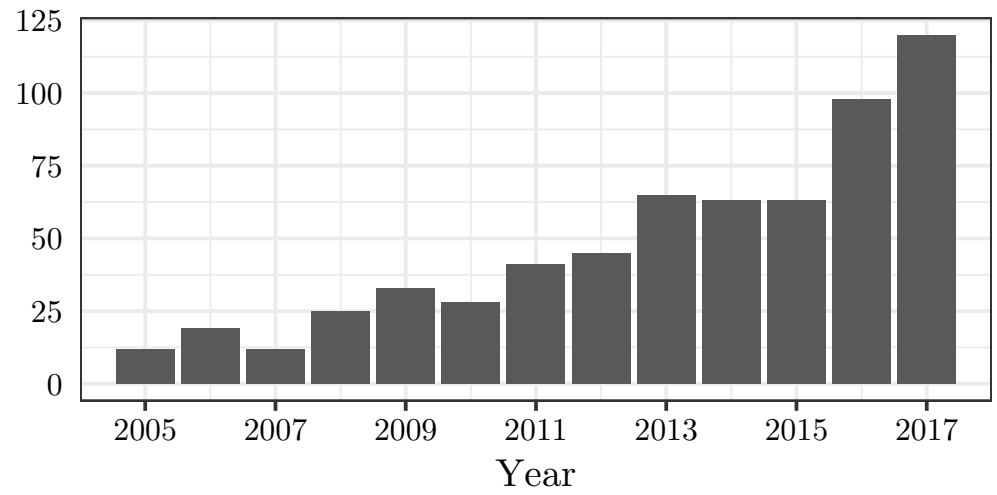

Fig. 3. Number of published studies over time.

Since, different authors might use different terms, not captured by our search phrases, to refer to the prediction target in question, we decided to also include all studies that cite the previously discovered 53 publications ("snowballing"). Applying the same exclusion criteria, we ended up with **83** more studies. Due to the close-knit nature of the process mining community, there is a considerable overlap between these 83 studies and the 53 studies that had been retrieved during the first search stage. Accordingly, a total of **110** publications were finally considered on the scope of our review. All papers retrieved at each search step can be found at https://goo.gl/kg6xZ1.

The remaining 110 papers were further assessed with respect to exclusion criteria:

- EX1 The study does not actually propose a predictive process monitoring *method*. With this criterion, we excluded position papers, as well as studies that, after a more thorough examination, turned out to be focusing on some research question other than predictive process monitoring. Furthermore, here we excluded survey papers and implementation papers that employ existing predictive methods rather than propose new ones.
- EX2 The study does not concern remaining time predictions. Common examples of other prediction targets that are considered irrelevant to this study are failure and error prediction, as well as next activity prediction. At the same time, prediction targets such as case completion time prediction and case duration prediction are inherently related to remaining time and therefore were also considered in our work. Additionally, this criterion does not eliminate studies that address the problem of predicting deadline violations in a boolean manner by setting a threshold on the predicted remaining time rather than by a direct classification.
- EX3 The study does not take an *event log* as input. In this way, we exclude methods that do not utilize at least the following essential parts of an event log: the case identifier, the timestamp and the event classes. For instance, we excluded methods that take as input numerical time series without considering the heterogeneity in the control flow. In particular, this is the case in manufacturing processes which are of linear nature (a process chain). The reason for excluding such studies is that the challenges when predicting for a set of cases of heterogenous length are different from those when predicting for linear processes. While methods designed for heterogenous processes are usually applicable to those of linear nature, it is not so vice versa. Moreover, the linear nature of a process makes it possible to apply other, more standard methods that may achieve better performance.

The application of the exclusion criteria resulted in **24** relevant studies which are described in detail in the following section.

## 4 ANALYSIS AND CLASSIFICATION OF METHODS

Driven by the research questions defined in Section 3.1, we identified the following dimensions to categorize and describe the relevant studies.

- Type of input data – RQ2
- Awareness of the underlying business process – RQ2
- Family of algorithms – RQ2
- Type of evaluation data (real-life or artificial log) and application domain (e.g., insurance, banking, healthcare) – RQ3.
- Type of implementation (standalone or plug-in, and tool accessibility) – RQ4

This information, summarized in Table 3, allows us to answer the first research question. In the remainder of this section, we proceed with surveying each main study method along the above classification dimensions.

### 4.1 Input data

As stipulated by *EX3* criterion, all surveyed proposals take as input an event log. Such a log contains at least a case identifier, an activity and a timestamp. In addition, many techniques leverage case and event attributes to make more accurate predictions. For example, in the pioneering predictive monitoring approach described in [54], the authors predict the remaining processing time of a trace using activity durations, their frequencies and various case attributes, such as the case priority. Many other approaches, e.g. [12], [36], [39] make use not only of case attributes but also of event attributes, while applying one or more kinds of sequence encoding. Furthermore, some approaches, e.g. [18] and [39], exploit contextual information, such as workload indicators, to take into account inter-case dependencies due to resource contention and data sharing. Finally, a group of works, e.g. [27] and [55] also leverage a process model in order to "replay" ongoing process cases on it. Such works treat remaining time as a cumulative indicator composed of cycle times of elementary process components.

### 4.2 Process awareness

Existing techniques can be categorized according to their process-awareness, i.e. whether or not the methodology exploits an explicit representation of a process model to make predictions. As can be seen from Table 3, nearly a half of the techniques are process-aware. Most of them construct a transition system from an event log using set, bag (multiset) or sequence abstractions of observed events. *State transition systems* are based on the idea that the process is composed of a set of consistent states and the movement between them [51]. Thus, a process behavior can be predicted if we know its current and future states.

Bolt and Sepúlveda [5] exploit query catalogs to store the information about the process behavior. Such catalogs are groups of partial traces (annotated with additional information about each partial trace) that have occurred in an event log, and are then used to estimate the remaining time of new executions of the process.

Also *queuing models* can be used for prediction because if a process follows a queuing context and queuing measures (e.g. arrival rate, departure rate) can be accurately estimated and fit the process actual execution, the movement of a queuing item can be reliably predicted. Queueing theory and regression-based techniques are combined for delay prediction in [40, 41].

Table 3. Overview of the 24 relevant studies resulting from the search (ordered by year and author).

| Study | Year | Input data | Process-aware? | Algorithm | Domain | Implementation |
|---|---|---|---|---|---|---|
| van Dongen et al. [54] | 2008 | event log<br>data | No | regression | public administration | ProM 5 |
| van der Aalst et al. [52] | 2011 | event log | Yes | transition system | public administration | ProM 5 |
| Folino et al. [18] | 2012 | event log<br>data<br>contextual information | No | clustering | logistics | ProM |
| Pika et al. [33] | 2012 | event log<br>data | No | stat analysis | financial | ProM |
| van der Spoel et al. [53] | 2012 | event log<br>data | Yes | process graph<br>regression | healthcare | n/a |
| Bevacqua et al. [4] | 2013 | event log<br>data | No | clustering<br>regression | losistics | ProM |
| Bolt and Sepúlveda [5] | 2013 | event log | Yes | stat analysis | telecom<br>simulated | n/a |
| Folino et al. [19] | 2013 | event log<br>data<br>contextual information | No | clustering | customer support | n/a |
| Pika et al. [34] | 2013 | event log<br>data | No | stat analysis | insurance | ProM |
| Rogge-Solti and Weske [37] | 2013 | event log<br>data | Yes | stoch Petri net | logistics<br>simulated | ProM |
| Ceci et al. [7] | 2014 | event log<br>data | Yes | sequence trees<br>regression | customer service<br>workflow management | ProM |
| Folino et al. [20] | 2014 | event log<br>data<br>contextual information | No | clustering<br>regression | logistics<br>software development | No |
| de Leoni et al. [11] | 2014 | event log<br>data<br>contextual information | No | regression | no validation | ProM |
| Polato et al. [35] | 2014 | event log<br>data | Yes | transition system<br>regression<br>classification | public administration | ProM |
| Senderovich et al. [40] | 2014 | event log | Yes | queueing theory<br>transition system | financial | n/a |
| Metzger et al. [27] | 2015 | event log<br>data<br>process model | Yes | neural network<br>constraint satisfaction<br>QoS aggregation | logistics | n/a |
| Rogge-Solti and Weske [38] | 2015 | event log<br>data | Yes | stoch Petri net | financial<br>logistics | ProM |
| Senderovich et al. [41] | 2015 | event log | Yes | queueing theory<br>transition system | financial<br>telecom | n/a |
| Cesario et al. [8] | 2016 | event log<br>data | Yes | clustering<br>regression | logistics | n/a |
| de Leoni et al. [12] | 2016 | event log<br>data<br>contextual information | No | regression | no validation | ProM |
| Polato et al. [36] | 2016 | event log<br>data | Yes | transition system<br>regression<br>classification | public administration<br>customer service | ProM |
| Senderovich et al. [39] | 2017 | event log<br>data<br>contextual information | No | regression | healthcare<br>manufacturing | standalone |
| Tax et al. [45] | 2017 | event log | No | neural network | customer support<br>public administration<br>financial | standalone |
| Verenich et al. [55] | 2017 | event log<br>data<br>process model | Yes | regression<br>classification<br>flow analysis | customer service<br>financial | standalone |

Furthermore, some process-aware approaches rely on stochastic Petri nets [37, 38] and process models in BPMN notation [55].

## 4.3 Family of algorithms

Non-process aware approaches typically rely on machine learning algorithms to make predictions. In particular, these algorithms take as input labeled training data in the form of feature vectors and the corresponding labels. In case of remaining time predictions, these labels are continuous. As such, various *regression* methods can be utilized, such as regression trees [11, 12] or ensemble of trees, i.e. random forest [53] and XGBoost [39].

An emerging family of algorithms for predictive monitoring are artificial neural networks. They consist of one layer of input units, one layer of output units, and multiple layers in-between which are referred to as hidden units. While traditional machine learning methods heavily depend on the choice of features on which they are applied, neural networks are capable of translating the data into a compact intermediate representation to aid a hand-crafted feature engineering process [1]. A feedforward network has been applied in [27] to predict deadline violations. More sophisticated architectures based on recurrent neural networks were explored in [45].

Other approaches apply trace clustering to group similar traces to fit a predictive model for each cluster. Then for any new running process case, predictions are made by using the predictor of the cluster it is estimated to belong to. Such approach is employed e.g., in [18] and [20].

Another range of proposals utilizes statistical methods without training an explicit machine learning model. For example, Pika et al. [33, 34] make predictions about time-related process risks by identifying and leveraging process risk indicators (e.g., abnormal activity execution time or multiple activity repetition) by applying statistical methods to event logs. The indicators are then combined by means of a prediction function, which allows for highlighting the possibility of transgressing deadlines. Conversely, Bolt and Sepúlveda [5] calculate remaining time based on the average time in the catalog which the partial trace belongs to, without taking into account distributions and confidence intervals.

Rogge-Solti and Weske [37] mine a stochastic Petri net from the event log to predict the remaining time of a case using arbitrary firing delays. The remaining time is evaluated based on the fact that there is an initial time distribution for a case to be executed. As inputs, the method receives the Petri net, the ongoing trace of the process instance up to current time, the current time and the number of simulation iterations. The algorithm returns the average of simulated completion times of each iteration. This approach is extended in [38] to exploit the elapsed time since the last observed event to make more accurate predictions.

Finally, Verenich et al. [55] propose a hybrid approach that relies on classification methods to predict routing probabilities for each decision point in a process model, regression methods to predict cycle times of future events, and flow analysis methods to calculate the total remaining time. A conceptually similar approach is proposed by Polato et al. [36] who build a transition system from an event log and enrich it with classification and regression models. Naive Bayes classifiers are used to estimate the probability of transition from one state to the other, while support vector regressors are used to predict the remaining time from the next state.

### 4.4 Evaluation data and domains

As reported in Table 3, most of the surveyed methods have been validated on at least one real-life event log. Some studies were additionally validated on simulated (synthetic) logs.

Importantly, many of the real-life logs are publicly available from the *4TU Center for Research Data*.[2] Among the methods that employ real-life logs, we observed a growing trend to use publicly-available logs, as opposed to private logs which hinder the reproducibility of the results due to not being accessible.

Concerning the application domains of the real-life logs, we noticed that most of them pertain to logistics, banking (7 studies each), public administration (5 studies) and customer service (3 studies) processes.

[2] https://data.4tu.nl/repository/collection:event_logs_real

### 4.5 Implementation

Nearly a half of the methods provide an implementation as a plug-in for the ProM framework.[3] The reason behind the popularity of ProM can be explained by its open-source and portable framework, which allows researchers to easily develop and test new algorithms. Also, ProM is historically the first process mining platform. Another three methods have a standalone implementation in Python. [4] Finally, 8 methods do not provide a publicly available implementation.

### 4.6 Predictive monitoring workflow

As indicated in Table 3, most predictive monitoring methods make use of machine learning algorithms based on regression, classification or neural networks. Such methods typically proceed in two phases: offline, to train a prediction model based on historical cases, and online, to make predictions on running process cases (Figure 4) [48].

In the offline phase, given an event log, case prefixes are extracted and filtered, e.g. to retain only prefixes up to a certain length, to create a prefix log (cf. Section 2). Next, "similar" prefixes are grouped into homogeneous *buckets*, e.g. based on process states or similarities among prefixes and prefixes from each bucket are *encoded* into feature vectors. Then feature vectors from each bucket are used to fit a machine learning model.

In the online phase, the actual predictions for running cases are made, by reusing the buckets, encoders and predictive models built in the offline phase. Specifically, given a running case and a set of buckets of historical prefixes, the correct bucket is first determined. Next, this information is used to encode the features of the running case. In the last step, a prediction is extracted from the encoded case using the pre-trained model corresponding to the determined bucket.

Similar observations can be made for non-machine learning-based methods. For example, in [52] first, a transition system is derived and annotated and then the actual predictions are calculated for running cases. In principle, this transition system akin to a predictive model can be mined in advance and used at runtime.

### 4.7 Primary and subsumed (related) studies

Among the papers that successfully passed both the inclusion and exclusion criteria, we determined *primary* studies that constitute an original contribution for the purposes of our benchmark, and *subsumed* studies that are similar to one of the primary studies and do not provide a substantial contribution with respect to it.

Specifically, a study is considered subsumed if:

- there exists a more recent and/or more extensive version of the study from the same authors (e.g. a conference paper is subsumed by an extended journal version), or
- it does not propose a substantial improvement/modification over a method that is documented in an earlier paper by other authors, or
- the main contribution of the paper is a case study or a tool implementation, rather than the predictive process monitoring method itself, and the method is described and/or evaluated more extensively in a more recent study by other authors.

This procedure resulted in **10** primary and **14** subsumed studies, as reported in Table 4. Some studies are subsumed by several primary studies. For instance, Metzger et al. [27] use a feedforward neural network which is subsumed by a more sophisticated recurrent neural network architecture extensively studied in [45]. At the same time, [27] describes a QoS technique which is subsumed by the flow analysis technique presented in [55].

[3] http://promtools.org

[4] https://www.python.org

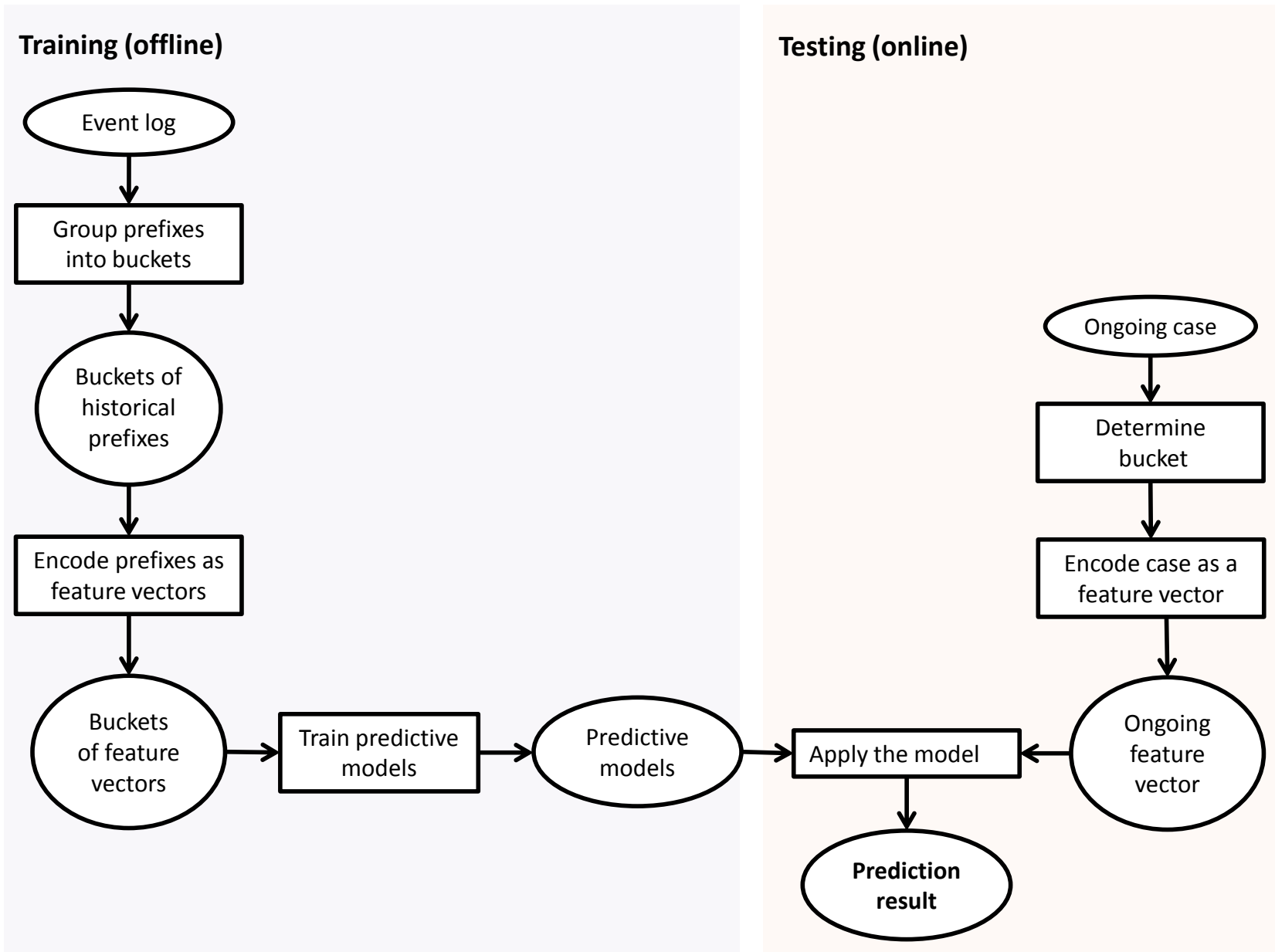

Fig. 4. Predictive process monitoring workflow.

Table 4. Primary and subsumed studies

| Primary study | Subsumed studies |
|---|---|
| van der Aalst et al. [52] | van Dongen et al. [54], Bolt and Sepúlveda [5] |
| Folino et al. [18] | Folino et al. [19, 20] |
| Rogge-Solti and Weske [38] | Rogge-Solti and Weske [37] |
| Senderovich et al. [41] | Senderovich et al. [40] |
| Cesario et al. [8] | Bevacqua et al. [4] |
| de Leoni et al. [12] | Pika et al. [33, 34], de Leoni et al. [11] |
| Polato et al. [36] | van der Spoel et al. [53], Polato et al. [35], Ceci et al. [7] |
| Senderovich et al. [39] | van der Spoel et al. [53] |
| Tax et al. [45] | Metzger et al. [27] |
| Verenich et al. [55] | Metzger et al. [27] |

## 5 METHODOLOGICAL FRAMEWORK

Assessing all the methods that resulted from the search would be infeasible due to the heterogeneous nature of the inputs required and the outputs produced. As such, we decided to *abstract* the details that are not inherent to the methods and focus on their core differences.

### 5.1 Prefix bucketing

While some machine learning-based predictive process monitoring approaches train a single predictor on the whole event log, others employ a multiple predictor approach by dividing the prefix traces in the historical log into several *buckets* and fitting a separate predictor for each bucket.

To this end, Teinemaa et al. [48] surveyed several bucketing methods out of which three have been utilized in the primary methods:

- Zero bucketing. All prefix traces are considered to be in the same bucket. As such, a single predictor is fit for all prefixes in the prefix log. This approach has been used in [12], [39] and [45].
- Prefix length bucketing. Each bucket contains the prefixes of a specific length. For example, the $n$-th bucket contains prefixes where at least $n$ events have been performed. One classifier is built for each possible prefix length. This approach has been used in [55].
- Cluster bucketing. Here, each bucket represents a cluster that results from applying a clustering algorithm on the encoded prefixes. One classifier is trained for each resulting cluster, considering only the historical prefixes that fall into that particular cluster. At runtime, the cluster of the running case is determined based on its similarity to each of the existing clusters and the corresponding classifier is applied. This approach has been used in [18] and [8]
- State bucketing. It is used in process-aware approaches where some kind of process representation, e.g. in the form of a transition system, is derived and a predictor is trained for each state, or decision point. At runtime, the current state of the running case is determined, and the respective predictor is used to make a prediction for the running case. This approach has been used in [36].

### 5.2 Prefix encoding

In order to train a machine learning model, all prefixes in a given bucket need to be represented, or *encoded* as fixed-size feature vectors. Case attributes are static and their number is fixed for a given process. Conversely, with each executed event, more information about the case becomes available. As such, the number of event attributes will increase over time. To address this issue, various sequence encoding techniques were proposed in related work summarized in [24] and refined in [48]. In the primary studies, the following encoding techniques can be found:

- Last state encoding. In this encoding method, only event attributes of the last $m$ events are considered. Therefore, the size of the feature vector is proportional to the number of event attributes and is fixed throughout the execution of a case. $m = 1$ is the most common choice used, e.g. in [36], although in principle higher $m$ values can also be used.
- Aggregation encoding. In contrast to the last state encoding, all events since the beginning of the case are considered. However, to keep the feature vector size constant, various aggregation functions are applied to the values taken by a specific attribute throughout the case lifetime. For numeric attributes, common aggregation functions are minimum, average, maximum and sum of observed values, while for categorical ones *count* is generally used, e.g. the number of times a specific activity has been executed, or the number of activities a specific resource has performed [12].
- Index-based encoding. Here for each position $n$ in a prefix, we concatenate the event $e_n$ occurring in that position and the value of each event attribute in that position $v_n^1, \ldots, v_n^k$, where $k$ is the total number of attributes of an event. This type of encoding is lossless, i.e. it is possible to recover the original prefix based on its feature vector. On the other hand, with longer prefixes, it significantly increases the dimensionality of the feature vectors and hinders the model training process. This approach has been used in [55].
- Tensor encoding. A tensor is a generalization of vectors and matrices to potentially higher dimensions [43]. Unlike conventional machine learning algorithms, tensor-based models do not require input to be encoded in a two-dimensional $n \times m$ form, where $n$ is the number of training instances and $m$ is the number of features. Conversely, they can take as input a three-dimensional tensor of shape $n \times t \times p$ , where $t$ is the number of events and $p$ is the number of event attributes, or features derived from each event. In other words, each prefix is represented as a matrix where rows correspond to events and columns to features for a given event. The data for each event is encoded ”as-is”. Case attributes are encoded as event attributes having the same value throughout the prefix. Hence, the encoding for LSTM is similar to the index-based encoding except for two differences: (i) case attributes are duplicated for each event, (ii) the feature vector is reshaped into a matrix.

To aid the explanation of the encoding types, Tables 5 – 7 provide examples of feature vectors derived from the event log in Table 1. Note that for the index-based encoding, each trace $\sigma$ in the event log produces only one training sample per bucket, while the other encodings produce as many samples as many prefixes can be derived from the original trace, i.e. up to $|\sigma|$.

Table 5. Feature vectors created from the log in Table 1 using last state encoding.

| Channel | Age | Activity_last | Time_last | Resource_last | Cost_last |
|---|---|---|---|---|---|
| Email | 37 | A | 0 | John | 15 |
| Email | 37 | B | 80 | Mark | 25 |
| Email | 37 | D | 180 | Mary | 10 |
| Email | 37 | F | 305 | Kate | 20 |
| Email | 37 | G | 350 | John | 20 |
| Email | 37 | H | 360 | Kate | 15 |
| Email | 52 | A | 0 | John | 25 |
| Email | 52 | D | 300 | Mary | 25 |
| Email | 52 | B | 57900 | Mark | 10 |
| Email | 52 | F | 58010 | Kate | 15 |

Table 6. Feature vectors created from the log in Table 1 using aggregated encoding.

| Channel | Age | Activ_A | Activ_B | Activ_D | Activ_F | Activ_G | Activ_H | Res_John | Res_Mark | Res_Mary | Res_Kate | sum_Time | sum_Cost | ... |
|---|---|---|---|---|---|---|---|---|---|---|---|---|---|---|
| Email | 37 | 1 | 0 | 0 | 0 | 0 | 0 | 1 | 0 | 0 | 0 | 0 | 15 | |
| Email | 37 | 1 | 1 | 0 | 0 | 0 | 0 | 1 | 1 | 0 | 0 | 80 | 40 | |
| Email | 37 | 1 | 1 | 1 | 0 | 0 | 0 | 1 | 1 | 1 | 0 | 180 | 50 | |
| Email | 37 | 1 | 1 | 1 | 1 | 0 | 0 | 1 | 1 | 1 | 1 | 305 | 70 | |
| Email | 37 | 1 | 1 | 1 | 1 | 1 | 0 | 2 | 1 | 1 | 1 | 350 | 90 | |
| Email | 37 | 1 | 1 | 1 | 1 | 1 | 1 | 2 | 1 | 1 | 2 | 360 | 105 | |
| Email | 52 | 1 | 0 | 0 | 0 | 0 | 0 | 1 | 0 | 0 | 0 | 0 | 25 | |
| Email | 52 | 1 | 0 | 1 | 0 | 0 | 0 | 1 | 0 | 1 | 0 | 300 | 50 | |
| Email | 52 | 1 | 1 | 1 | 0 | 0 | 0 | 1 | 1 | 1 | 0 | 57900 | 60 | |
| Email | 52 | 1 | 1 | 1 | 1 | 0 | 0 | 1 | 1 | 1 | 1 | 58010 | 75 | |

Table 7. Feature vectors created from the log in Table 1 using index-based encoding, buckets of length $n = 3$.

| Channel | Age | Activ_1 | Time_1 | Res_1 | Cost_1 | Activ_2 | Time_2 | Res_2 | Cost_2 | Activ_3 | Time_3 | Res_3 | Cost_3 |
|---|---|---|---|---|---|---|---|---|---|---|---|---|---|
| Email | 37 | A | 0 | John | 15 | B | 80 | Mark | 25 | D | 180 | Mary | 10 |
| Email | 52 | A | 0 | John | 25 | D | 300 | Mary | 25 | B | 57900 | Mark | 10 |

These three ”canonical” encodings can serve as a basis for various modifications thereof. For example, de Leoni et al. [12] proposed the possibility of combining last state and aggregation encodings.

While the encoding techniques stipulate how to incorporate *event* attributes in a feature vector, the inclusion of case attributes and inter-case metrics, such as the number of currently open cases, is rather straightforward, as their number is fixed throughout the case lifetime.

While last state and aggregation encodings can be combined with any of the bucketing methods described in Section 5.1, index-based encoding is commonly used with prefix-length bucketing, as the feature vector size depends on the trace length [24]. Nevertheless, two options have been proposed in related work to combine index-based encoding with other bucketing types:

- Fix the maximum prefix length and, for shorter prefixes, impute missing event attribute values with zeros or their historical averages. This approach is often referred to as *padding* in machine learning [42] and has been used in the context of predictive process monitoring in [46] and [29].
- Use the sliding window method to encode only recent (up-to window size $w$) history of the prefix. This approach has been proposed in [39].

### 5.3 Predictive algorithms

Machine learning-based predictive process monitoring methods have employed a variety of classification and regression algorithms, with the most popular choice being decision trees (e.g. [12, 20]). Although quite simple, decision trees have an advantage in terms of computation performance and interpretability of the results. Other choices include random forest [39, 55], support vector regression [36] and extreme gradient boosting [39].

A recent cross-benchmark of 13 state-of-the-art commonly used machine learning algorithms on a set of 165 publicly available classification problems [31] found that gradient tree boosting generally achieves better accuracy than random forest, while random forest is more accurate than decision trees. As a result, predictive process monitoring methods using predictive algorithms that are inherently more accurate will perform better.

In order to eliminate the bias associated with the usage of different predictors in related work, we decided to fix extreme gradient boosting (XGBoost) [9] as the main predictor across all the compared techniques. XGBoost is based on the theory of boosting, wherein the predictions of several "weak" learners (models whose predictions are slightly better than random guessing), are combined to produce a "strong" learner [21]. These "weak" learners are combined by following a gradient learning strategy. At the beginning of the calibration process, a "weak" learner is fit to the whole space of data, and then, a second learner is fit to the residuals of the first one. This process of fitting a model to the residuals of the previous one goes on until some stopping criterion is reached. Finally, the output of XGBoost is a weighted mean of the individual predictions of each weak learner. Regression trees are typically selected as "weak" learners [49].

Nevertheless, we set aside long short term memory (LSTM) neural networks as another predictor applied in the primary method [45]. We will apply a zero bucketing with LSTMs, i.e. train a single LSTM model for all prefixes, as this is the only bucketing type that has been used with LSTMs in the literature. As unlike other predictors, LSTMs do not require flattening the input data, we will naturally use tensor encoding with them. Similarly to the other predictors, to overcome the problem of feature vectors increasing with the prefix size, we fix the maximum number of events in the prefix (cf. Section 6.2.2) and for shorter prefixes, pad the data for missing events with zeros as in [29, 46].

### 5.4 Discussion

Summarizing the above observations, we devised a taxonomy of predictive monitoring techniques to be evaluated in our cross-benchmark (Figure 5). The taxonomy is framed upon a general classification of machine learning approaches into generative and discriminative ones (cf. Section 2.2).

The former correspond to process-aware predictive monitoring techniques, meaning that there is an assumption that an observed sequence is generated by some process that needs to be uncovered via probabilistic reasoning. The process can be represented via a state transition system [52], a Petri net [38] or a queueing model [41]. In our benchmark, we use implementations provided in [38, 52] as such, only making changes in the way the predictions are evaluated, in order to bring it in line with the other methods (cf. Section 6.2.2). The other primary method [41] had to be excluded due to the absence of a publicly available implementation.

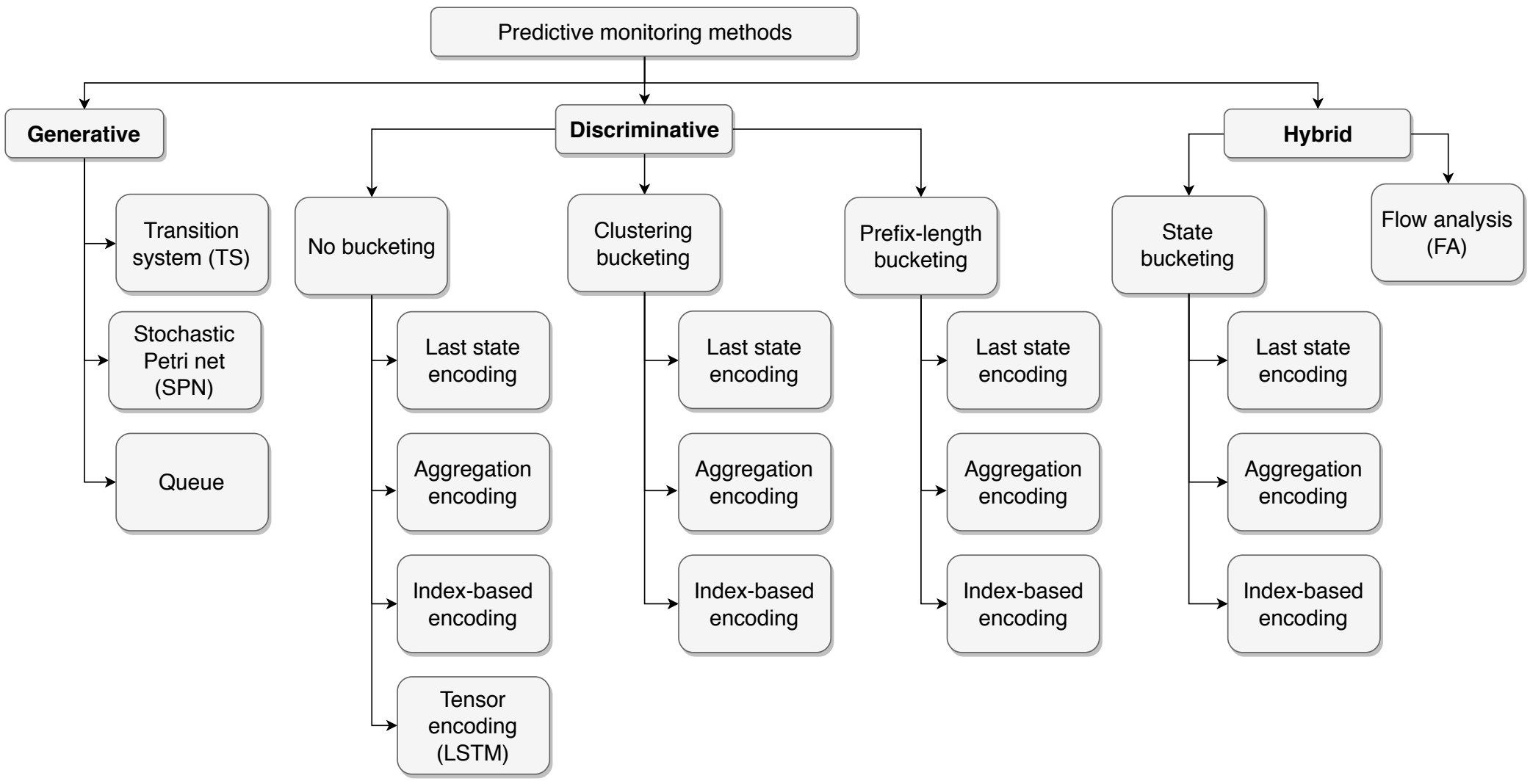

Fig. 5. Taxonomy of methods for predictive monitoring of remaining time.

Conversely, discriminative approaches are non-process-aware techniques that learn a direct mapping from inputs to the output via regression, without providing a model of how input sequences are actually generated. Having analyzed the discriminative studies, we have observed that they tend to mix different encoding types [12] or different bucketing types [18], while some combinations thereof have not been explored. As such, in our benchmark, we decided to evaluate all feasible combinations of canonical encoding and bucketing types,[5] including those that have not been used in any existing approach in the literature. In this way, we will be able to assess the benefits of each combination, while abstracting implementation nuances of each individual primary method.

Finally, a range of newer studies propose hybrid methods that combine generative and discriminative approaches [36, 55]. Such methods can generally be approximated with state bucketing that, for every process state, fits a model to predict the remaining time starting from that state. Alternatively, the flow analysis technique [55] provides a higher degree of granularity by splitting the remaining time into its integral components – cycle times of individual activities to be executed according to the process model.

## 6 BENCHMARK

Based on the methods identified in the previous subsection, we conducted an extensive benchmark to identify relative advantages and trade-offs in order to answer RQ5. In this section, we describe the datasets, the evaluation setup and metrics, and present the results of the benchmark.

[5] In order to combine index-based encoding with bucketings other than prefix length-based, we use zero padding.

### 6.1 Datasets

We conducted the experiments using 16 real-life event datasets. To ensure the reproducibility of the experiments, the logs we used are publicly available at the *4TU Center for Research Data*[6] as of March 2018, except for one log, which we obtained from the demonstration version of a software tool.

We excluded from the evaluation those logs that do not pertain to business processes (e.g., JUnit 4.12 Software Event Log and NASA Crew Exploration Vehicle). Such logs are usually not case-based or they contain only a few cases. Furthermore, we discarded the log that comes from the Environmental permit application process, as it is an earlier version of the BPIC 2015 event log from the same collection.

All the logs have been preprocessed beforehand to ensure the maximum achievable prediction accuracy. Firstly, since the training and validation procedures require having a complete history of each case, we remove incomplete cases, as well as cases that have been recorded not from their beginning. For example, in the Road traffic fines log[7], we consider traces where the last recorded event is *Send Fine* to be pending and therefore incomplete. Secondly, we perform some basic feature engineering. For instance, using event timestamps, we extract weekday, hour and duration since the previous event in the given case and since the beginning of the case (elapsed time) from each log. Additionally, for categorical variables with many possible values, if some values appear very rarely (in less than 10 cases), these rare values are marked as *other*. Finally, we check if there are any data attributes that are constant across all cases and events, or cross-correlated with other attributes and discard them.

Table 8 summarizes the basic characteristics of each resulting dataset, namely the number of complete cases, the ratio of distinct (unique) traces (DTR), the number of event classes, the average number of distinct events per trace (DER), average case length, i.e. the average number of events per case and its coefficient of variation (CV), average case duration (in days) and its CV, and the number of case and event attributes. The datasets possess a very diverse range of characteristics and originate from a variety of domains.

### 6.2 Experimental Setup

In this section, we describe our approach to split the event logs into training and test sets along the temporal dimension. Next, we provide a description of our evaluation criteria and the baselines to compare against. Finally, we discuss the hyperparameter optimization procedure employed in our benchmark.

*6.2.1 Data split.* In order to simulate a real-life situation where prediction models are trained using historical data and applied to running cases, we employ a so-called temporal split to divide the event log into train and test cases. Specifically, all cases in the log are ordered according to their start time and the first 80% are used to fit the models, while the remaining 20% are used to evaluate the prediction accuracy. In other words, the classifier is trained with all cases that started before a given date $T_1$ which would represent a current point in time in a real-life scenario, and the testing is done only on cases that start afterwards. Technically, cases that start before $T_1$ and are still running at $T_1$ should not be included in either set. However, to prevent the exclusion of a significant number of cases, in our experiments, we allow the two sets not to be completely temporally disjoint.

[6]https://data.4tu.nl/repository/collection:event_logs_real
[7]doi:10.4121/uuid:270fd440-1057-4fb9-89a9-b699b47990f5

Table 8. Statistics of the datasets used in the experiments.

| log | # cases | DTR | event classes | DER | mean case length | CV case length | mean case duration | CV case duration | # attributes (case+event) | Domain |
|---|---|---|---|---|---|---|---|---|---|---|
| bpic2011 | 1140 | 0.858 | 251 | 0.505 | 131.342 | 1.542 | 387.283 | 0.875 | 6+10 | HC |
| bpic2012a | 12007 | 0.001 | 10 | 1 | 4.493 | 0.404 | 7.437 | 1.563 | 1+4 | Fin |
| bpic2012o | 3487 | 0.002 | 7 | 1 | 4.562 | 0.126 | 15.048 | 0.606 | 1+4 | Fin |
| bpic2012w | 9650 | 0.235 | 6 | 0.532 | 7.501 | 0.97 | 11.401 | 1.115 | 1+5 | Fin |
| bpic2015_1 | 696 | 0.976 | 189 | 0.967 | 41.343 | 0.416 | 96.176 | 1.298 | 17+8 | PA |
| bpic2015_2 | 753 | 0.999 | 213 | 0.969 | 54.717 | 0.348 | 159.812 | 0.941 | 17+8 | PA |
| bpic2015_3 | 1328 | 0.968 | 231 | 0.975 | 43.289 | 0.355 | 62.631 | 1.555 | 18+8 | PA |
| bpic2015_4 | 577 | 0.998 | 179 | 0.97 | 42 | 0.346 | 110.835 | 0.87 | 15+8 | PA |
| bpic2015_5 | 1051 | 0.997 | 217 | 0.972 | 51.914 | 0.291 | 101.102 | 1.06 | 18+8 | PA |
| bpic2017 | 31509 | 0.207 | 26 | 0.878 | 17.826 | 0.32 | 21.851 | 0.593 | 3+15 | Fin |
| credit | 10035 | 0 | 8 | 1 | 8 | 0 | 0.948 | 0.899 | 0+7 | Fin |
| helpdesk | 3218 | 0.002 | 5 | 0.957 | 3.293 | 0.2 | 7.284 | 1.366 | 7+4 | CS |
| hospital | 59228 | 0 | 8 | 0.995 | 5.588 | 0.123 | 165.48 | 0.671 | 1+20 | HC |
| invoice | 5123 | 0.002 | 17 | 0.979 | 12.247 | 0.182 | 2.159 | 1.623 | 5+10 | FI |
| sepsis | 1035 | 0.076 | 6 | 0.995 | 5.001 | 0.288 | 0.029 | 1.966 | 23+10 | HC |
| traffic_fines | 150370 | 0.002 | 11 | 0.991 | 3.734 | 0.439 | 341.676 | 1.016 | 4+10 | PA |

*Domains:* HC – healthcare, Fin – financial, PA – public administration, CS – customer service

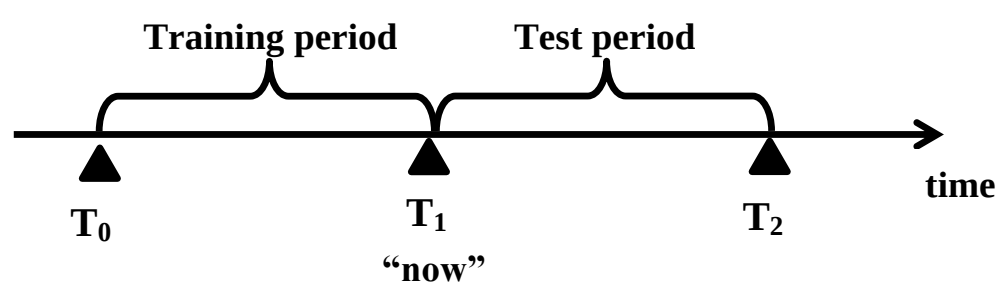

Fig. 6. Temporal split of the training and test sets.

*6.2.2 Evaluation metrics.* Two measures commonly employed to assess a predictive process monitoring technique are accuracy and earliness [14, 24, 47]. Indeed, in order to be useful, a prediction should be accurate and should be made early on to allow enough time to act upon.

**Accuracy.** To assess the accuracy of the prediction of continuous variables, well-known error metrics are Mean Absolute Error (MAE), Root Mean Square Error (RMSE) and Mean Percentage Error (MAPE) [22], where MAE is defined as the arithmetic mean of the prediction errors, RMSE as the square root of the average of the squared prediction errors, while MAPE measures error as the average of the unsigned percentage error. We observe that the value of remaining time tends to be highly varying across cases of the same process, sometimes with values on different orders of magnitude. RMSE would be very sensitive to such outliers. Furthermore, the remaining time can be very close to zero, especially near the end of the case, thus MAPE would be skewed in such situations. Hence, we use MAE to measure the error in predicting the remaining time. Formally,

$$MAE = \frac{1}{n} \sum_{i=1}^{n} |y_i - \hat{y}_i| \tag{2}$$

where $y_i \in \mathcal{Y} = \mathbb{R}$ is the actual value of a function in a given point and $\hat{y}_i \in \mathcal{Y} = \mathbb{R}$ is the predicted value.

**Earliness.** A common approach to measure the earliness of the predictions is to evaluate the accuracy of the models after each arrived event or at fixed time intervals. Naturally, uncertainty decreases as a case progresses towards its completion. Thus, the earlier we reach the desired level of accuracy, the better the technique is in terms of its earliness.

To measure earliness, we make predictions for prefixes $hd^k(\sigma)$ of traces $\sigma$ in the test set starting from $k = 1$. However, using all possible values of $k$ is coupled with several issues. Firstly, a large number of prefixes considerably increases the training time of the prediction models. Secondly, for a single model approach, the longer cases tend to produce much more prefixes than shorter ones and, therefore, the prediction model is biased towards the longer cases [48]. Finally, for a multiple model approach, if the distribution of case lengths has a long tail, for very long prefixes, there are not enough traces with that length, and the error measurements become unreliable. Consequently, we use prefixes of up to 20 events only in both training and test phase. If a case contains less than 20 events, we use all prefixes, except the last one, as predictions do not make sense when the case has completed. In other words, the input for our experiments is a filtered prefix log $L^* = \{hd^k(\sigma) : \sigma \in L, 1 \le k \le min(|\sigma| - 1, 20)\}$.

Inherently, there is often a trade-off between accuracy and earliness. As more events are executed, due to more information becoming available, the prediction accuracy tends to increase, while the earliness declines [39, 47]. As a result, we measure the performance of the models w.r.t. each dimension separately.

*6.2.3 Hyperparameter optimization.* Each prediction technique is characterized by model parameters and by hyperparameters [2]. While model parameters are learned during the training phase so as to fit the data, hyperparameters are set outside the training procedure and used for controlling how flexible the model is in fitting the data. For instance, the number of clusters $k$ in the k-means clustering procedure is a hyperparameter of the clustering technique. The impact of hyperparameter values on the accuracy of the predictions can be extremely high. Optimizing their value is therefore important, but, at the same time, optimal values depend on the specific dataset under examination [2].

A traditional way of performing hyperparameter optimization is *grid search*, which is an exhaustive search through a manually specified subset of the hyperparameter space of a learning algorithm. A grid search algorithm must be guided by some performance metric, typically measured by cross-validation on the training set or evaluation on a held-out validation set [28]. In our benchmark, to find the best set of hyperparameters, we perform grid search using five-fold cross-validation. For each combination of hyperparameters, we train a model based on the 80% of the original training set and evaluate its performance on the remaining validation set. The procedure is repeated five times, while measuring the validation performance over each split using the same metrics as for the test set. Then we select one combination of the parameters that achieves the best validation performance and retrain a model with these parameters, now using the whole training set.

For machine learning-based techniques, we use the implementation of XGBoost from the `scikit-learn` library [32] for `Python` which allows for a wide range of learning parameters as described in the work by Tianqi and Carlos [9]. Table 9 lists the most sensitive learning parameters that were optimized during the grid search. To achieve the optimal predictive power, these parameters are tuned for each combination of dataset, bucketing method, and sequence encoding method. For approaches that involve clustering, we use the k-means clustering algorithm, which is one of the most widely used clustering methods in the literature. K-means requires one to specify in advance the desired number of clusters $k$. We searched for the optimal value in the set $k \in \{2, 5, 10\}$.

In the case of the index-based bucketing method, an optimal configuration was chosen for each prefix length separately.

Table 9. Hyperparameters tuned via grid search.

| Parameter | Explanation | Search space |
|---|---|---|
| | *XGBoost* | |
| *n_estimators* | Number of decision trees ("weak" learners) in the ensemble | {250, 500} |
| *learning_rate* | Shrinks the contribution of each successive decision tree in the ensemble | {0.02, 0.04, 0.06} |
| *subsample* | Fraction of observations to be randomly sampled for each tree. | {0.5, 0.8} |
| *colsample_bytree* | Fraction of columns (features) to be randomly sampled for each tree. | {0.5, 0.8} |
| *max_depth* | Maximum tree depth for base learners | {3, 6} |
| | *LSTM* | |
| *units* | Number of neurons per hidden layer | {100, 200} |
| *n_layers* | Number of hidden layers | {1, 2, 3} |
| *batch* | Number of samples to be propagated | {8, 32} |
| *activation* | Activation function to use | $\{ReLU\}$ |
| *optimizer* | Weight optimizer | $\{RMSprop, Nadam\}$ |

For LSTMs, we used the recurrent neural network library [10], with *Tensorflow* backend. As with XGBoost, we performed tuning using grid search using the parameters specified in Table 9. These parameters and their values were chosen based on the results reported in [45]. Other hyperparameters were left to their defaults.

A similar procedure is performed for methods that do not train a machine learning model. For the method in [52], we vary the type of abstraction – set, bag of sequence – to create a transition system. For the method in [38], we vary the stochastic Petri net properties: (i) the kind of transition distribution – normal, gaussian kernel or histogram and (ii) memory semantics – global preselection or race with memory. We select the parameters that yield the best performance on the validation set and use them for the test set.

### 6.3 Evaluation results

Table 10 reports the prediction accuracy, averaged across all evaluated prefix lengths, together with its standard deviation. The averages are weighted by the relative frequency of prefixes with that prefix (i.e. longer prefixes get lower weights, since not all traces reach that length). In our experiences, we set an execution cap of 6 hours for each training configuration, so if some method did not finish within that time, the corresponding cell in the Table is empty. Furthermore, since the flow-analysis approach [55] cannot readily deal with unstructured process models, we only provide its results for the logs from which we were able to derive structured models.

Overall, we can see that in 13 out of 16 datasets, LSTM-based networks achieve the best accuracy, while flow-analysis and index-based encoding with no bucketing and prefix-length bucketing achieve the best results in one dataset each.

Figure 7 presents the prediction accuracy in terms of MAE, evaluated over different prefix lengths. Each evaluation point includes prefixes of exactly the given length. In other words, traces that are shorter than the required prefix are left out of the calculation. Therefore, the number of cases used for evaluation is monotonically decreasing when increasing the prefix length.

In most of the datasets, we see that the MAE decreases as cases progress. It is natural that the prediction task becomes trivial when cases are close to completion. However, for some datasets, the predictions become less accurate as the prefix length increases. This phenomenon is caused by the fact that these datasets contain some short traces for which it appears to be easy to predict the outcome. These short traces are not included in the later evaluation points, as they have already finished by that time. Therefore, we are left with longer traces only, which appear to be more challenging for the predictor, hence decreasing the total accuracy on larger prefix lengths.

Table 10. Weighted average MAE over all prefixes.

| Method | **bpic2011** | **bpic2012a** | **bpic2012o** | **bpic2012w** | **bpic2015_1** | **bpic2015_2** |
|---|---|---|---|---|---|---|
| TS [52] | 236.088 ± 9.98 | 8.363 ± 4.797 | 6.766 ± 2.909 | 7.505 ± 1.036 | 56.498 ± 8.341 | 118.293 ± 16.819 |
| LSTM [45] | **160.27 ± 24.325** | **3.772 ± 3.075** | 6.418 ± 2.768 | **6.344 ± 0.994** | 39.457 ± 5.708 | **61.62 ± 2.061** |
| SPN [38] | – | 7.693 ± 1.889 | 6.489 ± 2.562 | 8.538 ± 0.772 | 66.509 ± 17.131 | 81.114 ± 8.033 |
| FA [55] | – | 6.677 ± 3.72 | **5.95 ± 2.832** | 6.946 ± 1.057 | – | – |
| cluster_agg | 211.446 ± 5.941 | 6.739 ± 4.146 | 7.656 ± 3.534 | 7.18 ± 0.953 | 40.705 ± 1.824 | 68.185 ± 2.649 |
| cluster_index | 225.132 ± 5.212 | 6.743 ± 4.354 | 7.439 ± 3.436 | 7.074 ± 1.254 | 38.092 ± 2.988 | 66.957 ± 3.436 |
| cluster_last | 216.75 ± 4.338 | 6.728 ± 4.358 | 7.435 ± 3.412 | 7.061 ± 1.019 | 38.388 ± 3.478 | 62.781 ± 2.347 |
| prefix_agg | 211.401 ± 14.257 | 6.75 ± 4.452 | 7.79 ± 3.636 | 7.26 ± 0.935 | 46.765 ± 23.581 | 71.21 ± 8.893 |
| prefix_index | 227.288 ± 7.404 | 6.753 ± 4.45 | 7.472 ± 3.356 | 7.155 ± 0.942 | 37.525 ± 2.746 | 66.883 ± 3.756 |
| prefix_last | 219.781 ± 12.664 | 6.76 ± 4.429 | 7.441 ± 3.399 | 7.139 ± 0.851 | 37.975 ± 5.903 | 64.708 ± 5.749 |
| noBucket_agg | 200.466 ± 11.786 | 6.746 ± 3.899 | 7.744 ± 3.62 | 7.082 ± 1.02 | 35.962 ± 3.744 | 67.914 ± 2.467 |
| noBucket_index | 217.139 ± 13.991 | 6.768 ± 4.249 | 7.548 ± 3.367 | 6.982 ± 1.34 | **35.451 ± 2.499** | 65.505 ± 3.442 |
| noBucket_last | 208.711 ± 2.001 | 6.752 ± 4.15 | 7.51 ± 3.415 | 7.021 ± 1.099 | 37.442 ± 3.607 | 64.11 ± 2.332 |
| state_agg | 271.801 ± 14.676 | 6.756 ± 4.45 | 7.656 ± 3.534 | 7.465 ± 0.622 | 42.949 ± 2.725 | 68.768 ± 4.094 |
| state_index | – | 6.757 ± 4.453 | 7.439 ± 3.436 | 7.51 ± 0.585 | – | – |
| state_last | 271.595 ± 14.449 | 6.746 ± 4.446 | 7.435 ± 3.412 | 7.539 ± 0.554 | 42.946 ± 2.691 | 68.296 ± 3.762 |

| Method | **bpic2015_3** | **bpic2015_4** | **bpic2015_5** | **bpic2017** | **credit** | **helpdesk** |
|---|---|---|---|---|---|---|
| TS [52] | 26.412 ± 8.082 | 61.63 ± 5.413 | 67.699 ± 7.531 | 8.278 ± 2.468 | 0.382 ± 0.194 | 6.124 ± 2.6 |
| LSTM [45] | **19.682 ± 2.646** | **48.902 ± 1.527** | 52.405 ± 3.819 | **7.15 ± 2.635** | **0.062 ± 0.021** | **3.458 ± 2.542** |
| SPN [38] | 26.757 ± 10.378 | 51.202 ± 5.889 | – | 10.731 ± 0.369 | 0.385 ± 0.197 | 6.646 ± 1.225 |
| FA [55] | – | – | – | – | 0.075 ± 0.039 | 5.13 ± 2.092 |
| cluster_agg | 23.087 ± 3.226 | 51.555 ± 2.363 | 45.825 ± 3.028 | 7.479 ± 2.282 | 0.077 ± 0.036 | 4.179 ± 3.074 |
| cluster_index | 24.497 ± 1.887 | 56.113 ± 6.411 | 44.587 ± 4.378 | – | 0.075 ± 0.035 | 4.178 ± 3.043 |
| cluster_last | 22.544 ± 1.656 | 51.451 ± 4.189 | 46.433 ± 4.085 | 7.457 ± 2.359 | 0.076 ± 0.035 | 4.152 ± 3.053 |
| prefix_agg | 24.152 ± 2.785 | 53.568 ± 6.413 | 46.396 ± 2.466 | 7.525 ± 2.306 | 0.075 ± 0.034 | 4.175 ± 3.045 |
| prefix_index | 21.861 ± 3.292 | 50.452 ± 4.605 | **44.29 ± 3.669** | 7.421 ± 2.36 | 0.076 ± 0.035 | 4.262 ± 3.105 |
| prefix_last | 23.574 ± 3.778 | 53.053 ± 5.665 | 46.639 ± 3.718 | 7.482 ± 2.325 | 0.076 ± 0.034 | 4.242 ± 3.082 |
| noBucket_agg | 24.453 ± 3.577 | 54.89 ± 1.894 | 49.203 ± 1.833 | 7.437 ± 2.381 | 0.083 ± 0.033 | 4.252 ± 2.869 |
| noBucket_index | 23.025 ± 1.587 | 52.282 ± 1.182 | 50.153 ± 1.097 | – | 0.078 ± 0.034 | 4.253 ± 2.722 |
| noBucket_last | 25.15 ± 1.271 | 56.818 ± 1.729 | 49.027 ± 1.954 | 7.525 ± 2.244 | 0.082 ± 0.035 | 4.224 ± 2.814 |
| state_agg | 28.427 ± 9.844 | 49.318 ± 2.699 | 49.873 ± 2.658 | – | 0.077 ± 0.036 | 4.206 ± 3.092 |
| state_index | – | – | – | – | 0.079 ± 0.036 | 4.155 ± 3.023 |
| state_last | 27.826 ± 8.28 | 49.038 ± 2.498 | 49.556 ± 2.575 | 7.521 ± 2.341 | 0.078 ± 0.036 | 4.111 ± 3.026 |

| Method | **hospital** | **invoice** | **sepsis** | **traffic fines** |
|---|---|---|---|---|
| TS [52] | 46.491 ± 21.344 | 1.715 ± 0.891 | 0.019 ± 0.019 | 190.949 ± 15.447 |
| LSTM [45] | **36.258 ± 23.87** | **0.738 ± 0.266** | **0.009 ± 0.006** | **178.738 ± 89.019** |
| SPN [38] | 71.377 ± 29.082 | 1.646 ± 0.601 | 0.02 ± 0.032 | 193.807 ± 69.796 |
| FA [55] | 51.689 ± 14.945 | 1.224 ± 0.51 | – | 223.808 ± 14.859 |
| cluster_agg | 42.934 ± 26.136 | 1.048 ± 0.355 | 0.011 ± 0.006 | 210.322 ± 98.516 |
| cluster_index | – | 1.052 ± 0.404 | 0.011 ± 0.006 | 209.139 ± 98.417 |
| cluster_last | 48.589 ± 26.708 | 1.085 ± 0.389 | 0.011 ± 0.006 | 208.599 ± 99.549 |
| prefix_agg | 43.06 ± 25.884 | 1.03 ± 0.359 | 0.011 ± 0.006 | 212.614 ± 99.484 |
| prefix_index | 41.698 ± 25.944 | 1.041 ± 0.365 | 0.011 ± 0.006 | 209.085 ± 99.708 |
| prefix_last | 48.528 ± 26.714 | 1.057 ± 0.369 | 0.011 ± 0.006 | 209.304 ± 102.027 |
| noBucket_agg | 43.483 ± 25 | 1.186 ± 0.568 | 0.012 ± 0.005 | 211.017 ± 93.198 |
| noBucket_index | – | 1.053 ± 0.374 | 0.012 ± 0.005 | 208.879 ± 92.25 |
| noBucket_last | 50.496 ± 23.961 | 1.144 ± 0.466 | 0.012 ± 0.003 | 204.758 ± 93.399 |
| state_agg | 43.835 ± 25.984 | 1.044 ± 0.341 | 0.011 ± 0.006 | 211.439 ± 98.351 |
| state_index | 41.095 ± 26.499 | 1.051 ± 0.371 | 0.011 ± 0.006 | 210.408 ± 99.276 |
| state_last | 48.902 ± 27.001 | 1.086 ± 0.385 | 0.011 ± 0.006 | 209.206 ± 100.632 |

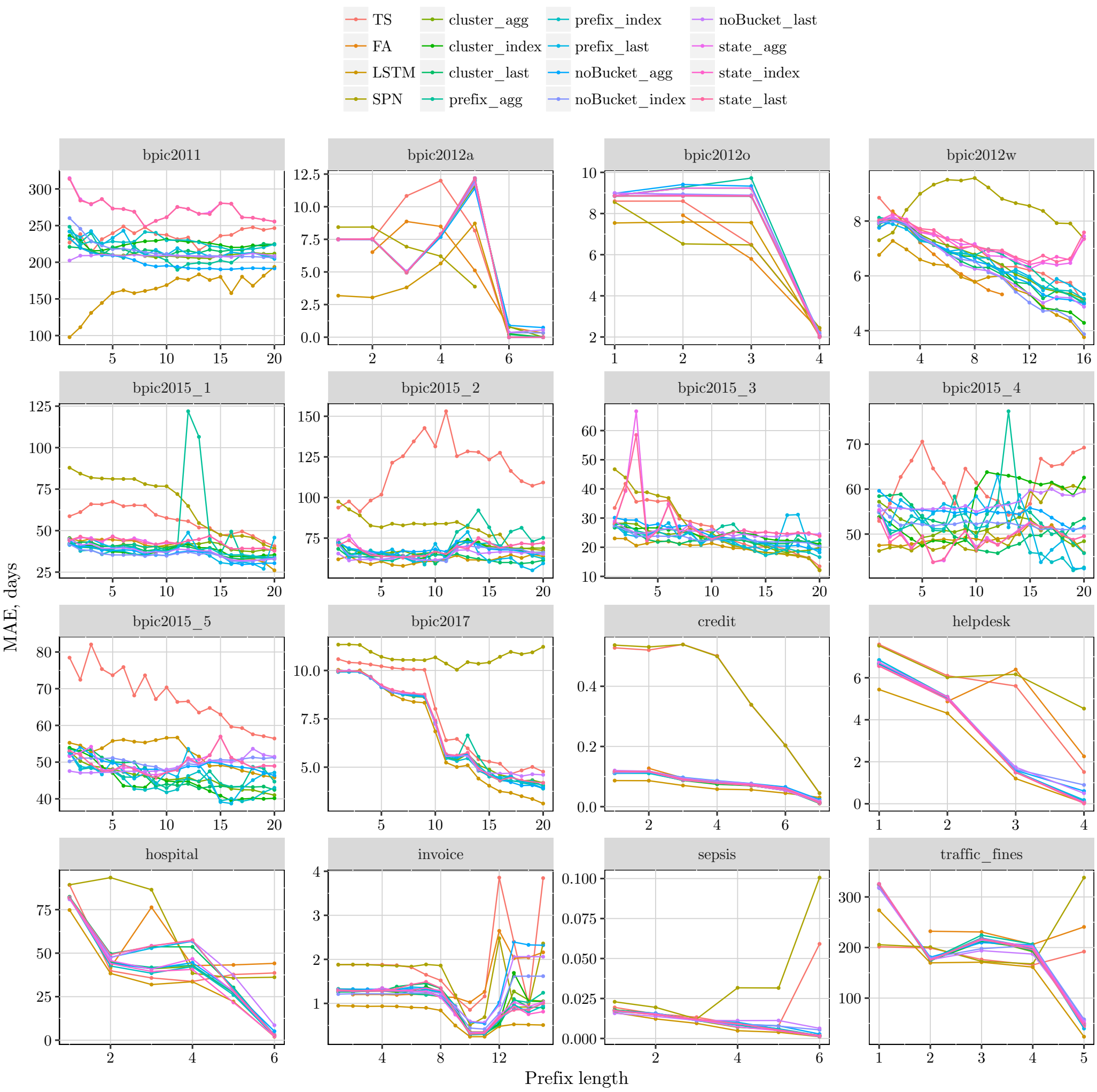

Fig. 7. Prediction accuracy (measured in terms of MAE) across different prefix lengths

As a simple bulk measure to compare the performance of the benchmarked techniques, we plot their mean rankings achieved across all datasets in Figure 8. Ties were resolved by assigning every tied element to the lowest rank. The rankings illustrate that LSTMs consistently outperform other machine-learning baselines in terms of accuracy (measured by MAE), while stochastic Petri nets and transition systems are usually the least accurate methods.

To complement the above observations, we also compare *aggregated* error values. These values need to be normalized, e.g. the by mean case duration, so that they are on a comparable scale. In order to do that, for each log, we divide the average MAE values and their standard deviations across all prefixes reported in Table 10 by the mean case duration for that log reported in Table 8. The results for each technique are illustrated in the boxplots in Figure 9, where each point represents the results for one of the 16 datasets. We can see that LSTM-based techniques have an average error of 40% of the mean case duration across all datasets. In contrast, transition systems on average incur a 59% error. Importantly, for LSTMs the accuracy varies between 0.07 and 0.56, while index-based encoding with prefix length bucketing, the method that on average achieves the second most accurate results, is more volatile and varies between 0.08 and 0.90 of the mean case duration.

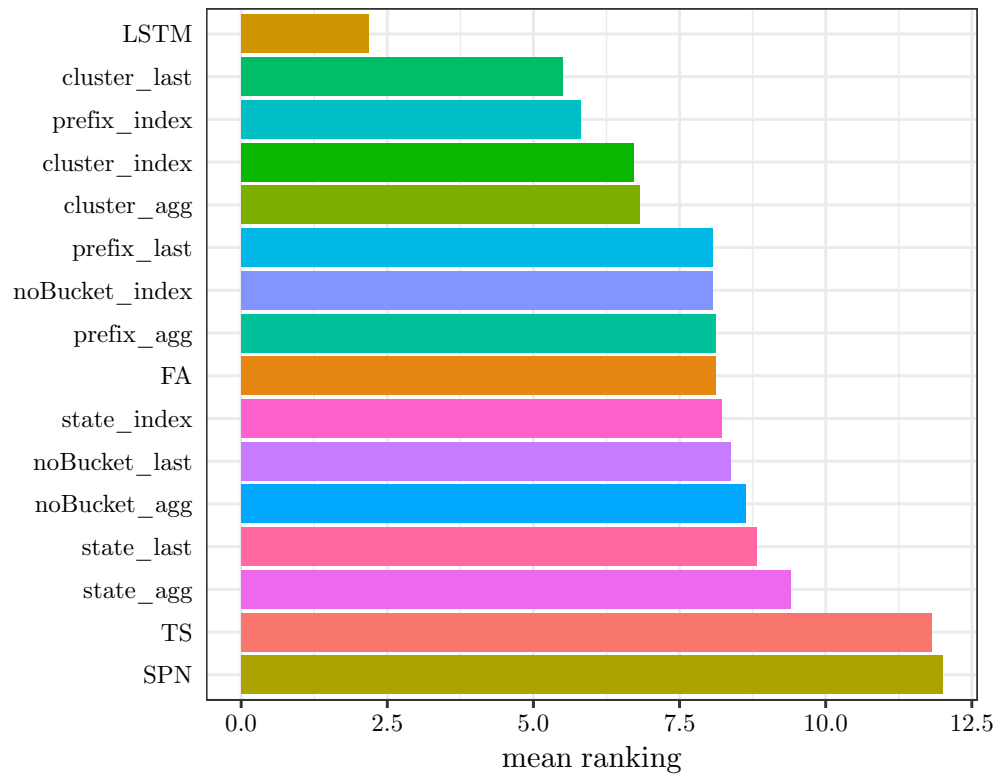

Fig. 8. Average ranking of the evaluated methods over all datasets. Error bars indicate the 95% confidence interval.

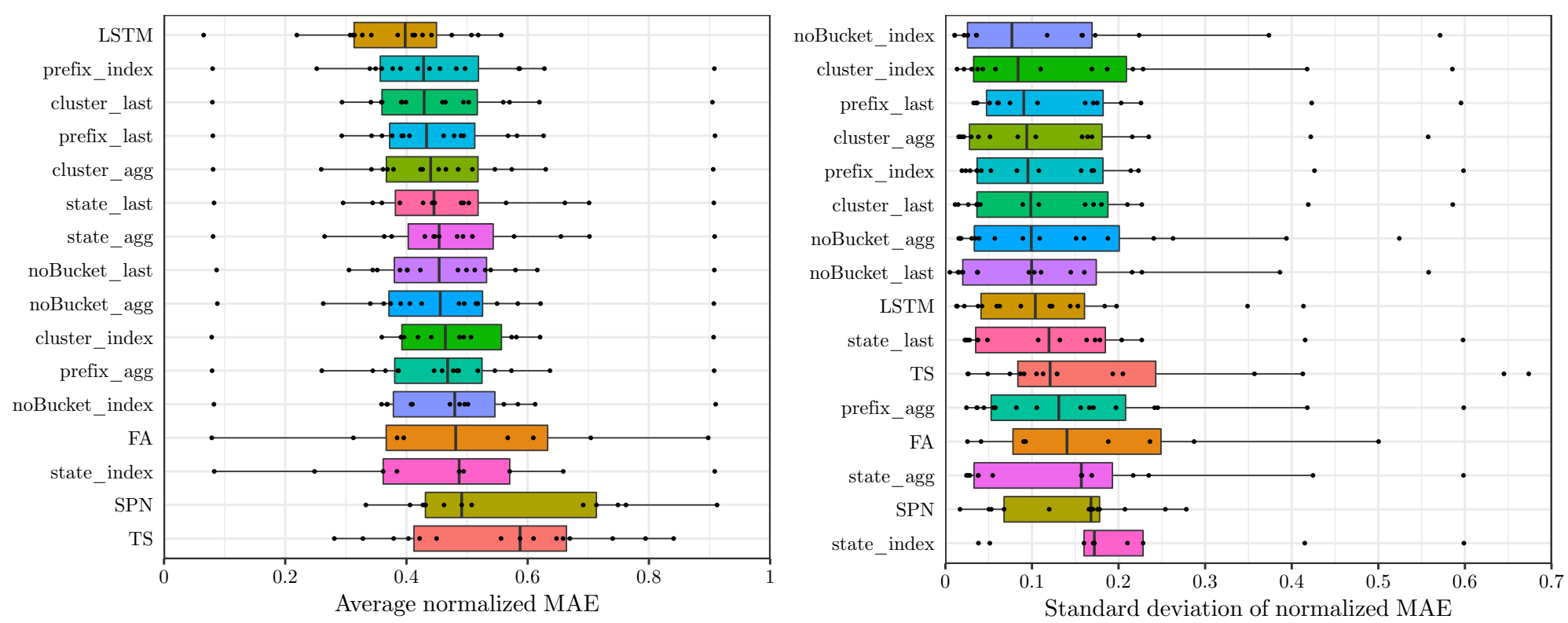

(a) (b)

Fig. 9. Average normalized MAE values (a) and their standard deviation (b) across case lifetime.

In order to assess the statistical significance of the observed differences in methods' performance across all datasets, we use the non-parametric Friedman test. The complete set of experiments indicate statistically significant differences according to this test ($p = 6.914 \times 10^{-8}$). Following the procedure suggested in the recent work on evaluating machine learning algorithms [13], in order to find which methods in particular differ from each other, we use the Nemenyi post-hoc test that compares all methods to each other.

Table 11. Post-hoc Nemenyi test of methods' rankings across all datasets.

| | cluster_agg | cluster_last | LSTM | noBucket_agg | noBucket_last | prefix_agg | prefix_index | prefix_last | state_last |
|---|---|---|---|---|---|---|---|---|---|
| cluster_last | 0.996 | | | | | | | | |
| LSTM | 0.061 | 0.464 | | | | | | | |
| noBucket_agg | 0.993 | 0.671 | **0.002** | | | | | | |
| noBucket_last | 0.993 | 0.671 | **0.002** | 1 | | | | | |
| prefix_agg | 0.997 | 0.748 | **0.003** | 1 | 1 | | | | |
| prefix_index | 0.999 | 1 | 0.366 | 0.766 | 0.766 | 0.831 | | | |
| prefix_last | 1 | 0.873 | **0.007** | 1 | 1 | 1 | 0.908 | | |
| state_last | 0.873 | 0.295 | **0.0001** | 1 | 1 | 1 | 0.424 | 0.996 | |
| TS | **0.043** | **0.002** | **0** | 0.444 | 0.444 | 0.366 | **0.003** | 0.262 | 0.783 |

Table 11 lists p-values of a pairwise post-hoc analysis. Since the test requires complete information for all pairwise comparisons, we included only 10 methods for which we have results on all 16 datasets. For most pairs, the null hypothesis that their performance is similar can not be rejected. However, the test underlines the impressive performance of LSTM, which significantly outperforms most of the other methods at the $p < 0.05$ level.

While on average most combinations of bucketing and encoding methods provide more or less similar levels of accuracy, we can observe differences for individual datasets. For example, in the hospital dataset, it is clear that clustering with aggregation encoding is better than with clustering with last state encoding. Arguably, aggregating knowledge from all events performed so far provides much more signal than using raw attributes of the latest event.

In order to explain differences in performance of various bucketing and encoding combinations, we try to correlate the characteristics of event logs (Table 8) with the type of bucketing/encoding that achieves the best accuracy on that log. One can notice that if case in the log are very heterogeneous in terms of case length, i.e. the coefficient of variation of case length is high enough, it is more beneficial to assign all traces to the same bucket. This can be observed in *bpic2012w* and *bpic2011* event logs, where standard deviation of case length is close to or exceeds mean case length. Furthermore, if cases in the log are short (e.g. in *helpdesk*, *traffic_fines*, *bpic2012a*, *bpic2012o*) or very distinctive from each other (e.g. in *bpic2015_2*), last state encoding tends to capture the most signal. Notably, in the aforementioned logs, the index-based encoding, although lossless, is not optimal. This suggests that in these datasets, combining the knowledge from all events performed so far provides more signal for remaining time prediction than the order of events. However, standard classifiers like XGBoost are not able to learn such higher-level features, unlike LSTMs, which is why in some situations even the simple aggregations outperform the index-based encoding.

## 7 THREATS TO VALIDITY

One of the threats to the validity of this study relates to the potential selection bias in the literature review. To minimize this, we described our systematic literature review procedure on a level of detail that is sufficient to replicate the search. However, in time the search and ranking algorithms of the used academic database (Google Scholar) might be updated and return different results. Another potential source of bias is the subjectivity when applying inclusion and exclusion criteria, as well as when determining the primary and subsumed studies. In order to alleviate this issue, all the included papers were collected in a publicly available spreadsheet, together with decisions and reasons about excluding them from the study. Moreover, each paper was independently assessed against the inclusion and exclusion criteria by two authors, and inconsistencies were resolved with the mediation of a third author.

Another threat to validity is related to the comprehensiveness of the conducted experiments. In particular, only one representative setting of each technique was used. Namely, only one machine learning algorithm (XGBoost) and one clustering method (k-means) were tested over all relevant methods. It is possible that there exists, for example, a combination of an untested clustering technique and a predictor that outperforms the settings used in this study. Furthermore, the generalizability of the findings is to some extent limited by the fact that the experiments were performed only on 16 event logs. Although these are all real-life event logs from different application fields that exhibit different characteristics, it may possible that the results would be different using other datasets or different log preprocessing techniques for the same datasets. In order to mitigate these threats, we built an open-source software framework which allows the full replication of the experiments, and made this tool publicly available. Moreover, additional datasets, as well as new sequence classification and encoding methods can be plugged in, so that the framework can be used for future experiments.

## 8 CONCLUSION

This study provided a survey, a comparative analysis and an evaluation of process monitoring techniques to predict the remaining time. The relevant existing studies were identified through a systematic literature review, which retrieved 24 novel studies dealing with the problem of remaining time prediction. Out of these, ten were considered to contain a distinct contribution (primary studies). Through further analysis of the primary studies, a taxonomy was proposed based on three main aspects, the type of input data required, process awareness and the family of algorithms employed.

It has been found that most studies, 7 out of 10, employ machine learning algorithms to train predictive models and apply them at runtime. These methods were further broken down into categories depending on how they divide the input traces into homogeneous buckets and how these traces are encoded into feature vectors. Each such combination was evaluated as a separate method for evaluation purposes. In addition, we evaluated long-short term memory (LSTM) networks and flow analysis as separate methods due to their peculiar properties. Finally, a comparative evaluation of the 16 identified techniques was performed using a unified experimental set-up and 16 real-life event logs. To ensure reproducibility of the results, all the selected techniques were implemented as a publicly available framework.

The results of the benchmark show that the most accurate results are obtained using LSTM networks, possibly owing to their ability to automatically learn relevant features from trace prefixes. Furthermore, we discovered that on average, there is no statistically significant difference between different canonical combinations of bucketing and encodings. As such, when simpler algorithms perform on par with a more complex one, it is often preferable to choose the simpler of the two. However, we showed that due to the properties exhibited by specific event logs, some combinations may achieve more accurate results.

All the source code required to reproduce the reported evaluation is available at https://github.com/verenich/time-prediction-benchmark. All the datasets employed are publicly available via the links provided in Section 6.1.

## ACKNOWLEDGMENTS

This research is partly funded by the Australian Research Council (grant DP180102839) and the Estonian Research Council (grant IUT20-55). Computational resources and services used in this work were provided by the HPC and Research Support Group, Queensland University of Technology, Brisbane, Australia.

## REFERENCES

[1] Yoshua Bengio, Aaron C. Courville, and Pascal Vincent. 2013. Representation Learning: A Review and New Perspectives. *IEEE Trans. Pattern Anal. Mach. Intell.* 35, 8 (2013), 1798–1828. https://doi.org/10.1109/TPAMI.2013.50

[2] James Bergstra, Rémi Bardenet, Yoshua Bengio, and Balázs Kégl. 2011. Algorithms for Hyper-Parameter Optimization. In *Advances in Neural Information Processing Systems 24: 25th Annual Conference on Neural Information Processing Systems 2011. Proceedings of a meeting held 12-14 December 2011, Granada, Spain.* 2546–2554. http://papers.nips.cc/paper/4443-algorithms-for-hyper-parameter-optimization

[3] JM Bernardo, MJ Bayarri, JO Berger, AP Dawid, D Heckerman, AFM Smith, and M West. 2007. Generative or discriminative? getting the best of both worlds. *Bayesian statistics* 8, 3 (2007), 3–24.

[4] Antonio Bevacqua, Marco Carnuccio, Francesco Folino, Massimo Guarascio, and Luigi Pontieri. 2013. A Data-Driven Prediction Framework for Analyzing and Monitoring Business Process Performances. In *Enterprise Information Systems - 15th International Conference, ICEIS 2013, Angers, France, July 4-7, 2013.* 100–117. https://doi.org/10.1007/978-3-319-09492-2_7

[5] Alfredo Bolt and Marcos Sepúlveda. 2013. Process Remaining Time Prediction Using Query Catalogs. In *Business Process Management Workshops - BPM 2013 International Workshops, Beijing, China, August 26, 2013, Revised Papers.* 54–65. https://doi.org/10.1007/978-3-319-06257-0_5

[6] Malú Castellanos, Fabio Casati, Umeshwar Dayal, and Ming-Chien Shan. 2004. A Comprehensive and Automated Approach to Intelligent Business Processes Execution Analysis. *Distributed and Parallel Databases* 16, 3 (2004), 239–273. https://doi.org/10.1023/B:DAPD.0000031635.88567.65

[7] Michelangelo Ceci, Pasqua Fabiana Lanotte, Fabio Fumarola, Dario Pietro Cavallo, and Donato Malerba. 2014. Completion Time and Next Activity Prediction of Processes Using Sequential Pattern Mining. In *Discovery Science - 17th International Conference, DS 2014, Bled, Slovenia, October 8-10, 2014. Proceedings.* 49–61. https://doi.org/10.1007/978-3-319-11812-3_5

[8] Eugenio Cesario, Francesco Folino, Massimo Guarascio, and Luigi Pontieri. 2016. A Cloud-Based Prediction Framework for Analyzing Business Process Performances. In *Availability, Reliability, and Security in Information Systems - IFIP WG 8.4, 8.9, TC 5 International Cross-Domain Conference, CD-ARES 2016, and Workshop on Privacy Aware Machine Learning for Health Data Science, PAML 2016, Salzburg, Austria, August 31 - September 2, 2016, Proceedings.* 63–80. https://doi.org/10.1007/978-3-319-45507-5_5

[9] Tianqi Chen and Carlos Guestrin. 2016. XGBoost: A Scalable Tree Boosting System *(Proceedings of the 22nd ACM SIGKDD International Conference on Knowledge Discovery and Data Mining, San Francisco, CA, USA, August 13-17, 2016).* 785–794. https://doi.org/10.1145/2939672.2939785

[10] François Chollet et al. 2015. Keras. https://github.com/fchollet/keras.

[11] Massimiliano de Leoni, Wil M. P. van der Aalst, and Marcus Dees. 2014. A General Framework for Correlating Business Process Characteristics *(BPM)*. 250–266.

[12] Massimiliano de Leoni, Wil M. P. van der Aalst, and Marcus Dees. 2016. A general process mining framework for correlating, predicting and clustering dynamic behavior based on event logs. *Information Systems* 56 (2016), 235–257.

[13] Janez Demsar. 2006. Statistical Comparisons of Classifiers over Multiple Data Sets. *Journal of Machine Learning Research* 7 (2006), 1–30. http://www.jmlr.org/papers/v7/demsar06a.html

[14] Chiara Di Francescomarino, Marlon Dumas, Fabrizio M Maggi, and Irene Teinemaa. 2017. Clustering-based predictive process monitoring. *IEEE Transactions on Services Computing* (2017).

[15] Marlon Dumas, Marcello La Rosa, Jan Mendling, and Hajo A. Reijers. 2018. *Fundamentals of Business Process Management, Second Edition.* Springer. https://doi.org/10.1007/978-3-662-56509-4

[16] Joerg Evermann, Jana-Rebecca Rehse, and Peter Fettke. 2016. A Deep Learning Approach for Predicting Process Behaviour at Runtime *(Business Process Management Workshops - BPM 2016 International Workshops, Rio de Janeiro, Brazil, September 19, 2016, Revised Papers).* 327–338. https://doi.org/10.1007/978-3-319-58457-7_24

[17] Usama M. Fayyad, Gregory Piatetsky-Shapiro, Padhraic Smyth, and Ramasamy Uthurusamy (Eds.). 1996. *Advances in Knowledge Discovery and Data Mining.* AAAI/MIT Press.

[18] Francesco Folino, Massimo Guarascio, and Luigi Pontieri. 2012. Discovering Context-Aware Models for Predicting Business Process Performances. In *On the Move to Meaningful Internet Systems: OTM 2012, Confederated International Conferences: CoopIS, DOA-SVI, and ODBASE 2012, Rome, Italy, September 10-14, 2012. Proceedings, Part I.* 287–304. https://doi.org/10.1007/978-3-642-33606-5_18

[19] Francesco Folino, Massimo Guarascio, and Luigi Pontieri. 2013. Discovering High-Level Performance Models for Ticket Resolution Processes. In *On the Move to Meaningful Internet Systems: OTM 2013 Conferences - Confederated International Conferences: CoopIS, DOA-Trusted Cloud, and ODBASE 2013, Graz, Austria, September 9-13, 2013. Proceedings.* 275–282. https://doi.org/10.1007/978-3-642-41030-7_18

[20] Francesco Folino, Massimo Guarascio, and Luigi Pontieri. 2014. Mining Predictive Process Models out of Low-level Multidimensional Logs. In *Advanced Information Systems Engineering - 26th International Conference, CAiSE 2014, Thessaloniki, Greece, June 16-20, 2014. Proceedings.* 533–547. https://doi.org/10.1007/978-3-319-07881-6_36

[21] Trevor Hastie, Robert Tibshirani, and Jerome H. Friedman. 2009. *The elements of statistical learning: data mining, inference, and prediction, 2nd Edition.* Springer. http://www.worldcat.org/oclc/300478243

[22] Rob J Hyndman and Anne B Koehler. 2006. Another look at measures of forecast accuracy. *International Journal of Forecasting* 22, 4 (2006), 679–688.

[23] Barbara Kitchenham. 2004. Procedures for performing systematic reviews. *Keele, UK, Keele University* 33, 2004 (2004), 1–26.

[24] Anna Leontjeva, Raffaele Conforti, Chiara Di Francescomarino, Marlon Dumas, and Fabrizio Maria Maggi. 2015. Complex Symbolic Sequence Encodings for Predictive Monitoring of Business Processes. In *Business Process Management - 13th International Conference*. 297–313.

[25] Fabrizio Maria Maggi, Chiara Di Francescomarino, Marlon Dumas, and Chiara Ghidini. 2014. Predictive monitoring of business processes *(CAiSE)*. . Springer, 457–472.

[26] A. E. Márquez-Chamorro, M. Resinas, and A. Ruiz-Corts. 2017. Predictive monitoring of business processes: a survey. *IEEE Transactions on Services Computing* PP, 99 (2017), 1–1. https://doi.org/10.1109/TSC.2017.2772256

[27] Andreas Metzger, Philipp Leitner, Dragan Ivanovic, Eric Schmieders, Rod Franklin, Manuel Carro, Schahram Dustdar, and Klaus Pohl. 2015. Comparing and Combining Predictive Business Process Monitoring Techniques. *IEEE Trans. Systems, Man, and Cybernetics: Systems* 45, 2 (2015), 276–290. https://doi.org/10.1109/TSMC.2014.2347265

[28] Tom M. Mitchell. 1997. *Machine learning*. McGraw-Hill.

[29] Nicolò Navarin, Beatrice Vincenzi, Mirko Polato, and Alessandro Sperduti. 2017. LSTM networks for data-aware remaining time prediction of business process instances. In *2017 IEEE Symposium Series on Computational Intelligence, SSCI 2017, Honolulu, HI, USA, November 27 - Dec. 1, 2017*. 1–7. https://doi.org/10.1109/SSCI.2017.8285184

[30] Andrew Y. Ng and Michael I. Jordan. 2001. On Discriminative vs. Generative Classifiers: A comparison of logistic regression and naive Bayes. In *Advances in Neural Information Processing Systems 14 [Neural Information Processing Systems: Natural and Synthetic, NIPS 2001, December 3-8, 2001, Vancouver, British Columbia, Canada]*. 841–848. http://papers.nips.cc/paper/2020-on-discriminative-vs-generative-classifiers-a-comparison-of-logistic-regression-and-naive-bayes

[31] Randal S. Olson, William La Cava, Zairah Mustahsan, Akshay Varik, and Jason H. Moore. 2017. Data-driven Advice for Applying Machine Learning to Bioinformatics Problems. *Arxiv* abs/1702.01780 (2017). http://arxiv.org/abs/1702.01780

[32] F. Pedregosa, G. Varoquaux, A. Gramfort, V. Michel, B. Thirion, O. Grisel, M. Blondel, P. Prettenhofer, R. Weiss, V. Dubourg, J. Vanderplas, A. Passos, D. Cournapeau, M. Brucher, M. Perrot, and E. Duchesnay. 2011. Scikit-learn: Machine Learning in Python. *Journal of Machine Learning Research* 12 (2011), 2825–2830.

[33] Anastasiia Pika, Wil M. P. van der Aalst, Colin J. Fidge, Arthur H. M. ter Hofstede, and Moe Thandar Wynn. 2012. Predicting Deadline Transgressions Using Event Logs. In *Business Process Management Workshops - BPM 2012 International Workshops, Tallinn, Estonia, September 3, 2012. Revised Papers*. 211–216. https://doi.org/10.1007/978-3-642-36285-9_22

[34] Anastasiia Pika, Wil M. P. van der Aalst, Colin J. Fidge, Arthur H. M. ter Hofstede, and Moe Thandar Wynn. 2013. Profiling Event Logs to Configure Risk Indicators for Process Delays. In *Advanced Information Systems Engineering - 25th International Conference, CAiSE 2013, Valencia, Spain, June 17-21, 2013. Proceedings*. 465–481. https://doi.org/10.1007/978-3-642-38709-8_30

[35] Mirko Polato, Alessandro Sperduti, Andrea Burattin, and Massimiliano de Leoni. 2014. Data-aware remaining time prediction of business process instances *(2014 International Joint Conference on Neural Networks, IJCNN 2014)*. 816–823.

[36] Mirko Polato, Alessandro Sperduti, Andrea Burattin, and Massimiliano de Leoni. 2016. Time and Activity Sequence Prediction of Business Process Instances. *CoRR* abs/1602.07566 (2016). arXiv:1602.07566 http://arxiv.org/abs/1602.07566

[37] Andreas Rogge-Solti and Mathias Weske. 2013. Prediction of Remaining Service Execution Time Using Stochastic Petri Nets with Arbitrary Firing Delays. In *International Conference on Service-Oriented Computing (ICSOC)*. Springer, 389–403.

[38] Andreas Rogge-Solti and Mathias Weske. 2015. Prediction of business process durations using non-Markovian stochastic Petri nets. *Information Systems* 54 (2015), 1–14. https://doi.org/10.1016/j.is.2015.04.004

[39] Arik Senderovich, Chiara Di Francescomarino, Chiara Ghidini, Kerwin Jorbina, and Fabrizio Maria Maggi. 2017. Intra and Inter-case Features in Predictive Process Monitoring: A Tale of Two Dimensions. In *Business Process Management - 15th International Conference, BPM 2017, Barcelona, Spain, September 10-15, 2017, Proceedings*. 306–323. https://doi.org/10.1007/978-3-319-65000-5_18

[40] Arik Senderovich, Matthias Weidlich, Avigdor Gal, and Avishai Mandelbaum. 2014. Queue Mining - Predicting Delays in Service Processes *(CAiSE)*. 42–57.

[41] Arik Senderovich, Matthias Weidlich, Avigdor Gal, and Avishai Mandelbaum. 2015. Queue mining for delay prediction in multi-class service processes. *Inf. Syst.* 53 (2015), 278–295. https://doi.org/10.1016/j.is.2015.03.010

[42] Xingjian Shi, Zhourong Chen, Hao Wang, Dit-Yan Yeung, Wai-Kin Wong, and Wang-chun Woo. 2015. Convolutional LSTM Network: A Machine Learning Approach for Precipitation Nowcasting. In *Advances in Neural Information Processing Systems 28: Annual Conference on Neural Information Processing Systems 2015, December 7-12, 2015, Montreal, Quebec, Canada*. 802–810. http://papers.nips.cc/paper/5955-convolutional-lstm-network-a-machine-learning-approach-for-precipitation-nowcasting

[43] Jimeng Sun, Dacheng Tao, and Christos Faloutsos. 2006. Beyond streams and graphs: dynamic tensor analysis. In *Proceedings of the Twelfth ACM SIGKDD International Conference on Knowledge Discovery and Data Mining, Philadelphia, PA, USA, August 20-23, 2006*. 374–383. https://doi.org/10.1145/1150402.1150445

[44] P.N. Tan, M. Steinbach, A. Karpatne, and V. Kumar. 2013. *Introduction to Data Mining*. Pearson Education. https://books.google.com.au/books?id=_ZQ4MQEACAAJ

[45] Niek Tax, Ilya Verenich, Marcello La Rosa, and Marlon Dumas. 2017. Predictive Business Process Monitoring with LSTM Neural Networks *(Advanced Information Systems Engineering - 29th International Conference, CAiSE 2017, Essen, Germany, June 12-16, 2017, Proceedings)*. 477–492. https://doi.org/10.1007/978-3-319-59536-8_30

[46] Irene Teinemaa, Marlon Dumas, Anna Leontjeva, and Fabrizio Maria Maggi. 2017. Temporal Stability in Predictive Process Monitoring. *CoRR* abs/1712.04165 (2017). arXiv:1712.04165 http://arxiv.org/abs/1712.04165

[47] Irene Teinemaa, Marlon Dumas, Fabrizio Maria Maggi, and Chiara Di Francescomarino. 2016. Predictive Business Process Monitoring with Structured and Unstructured Data *(Business Process Management - 14th International Conference, BPM 2016, Rio de Janeiro, Brazil, September 18-22, 2016)*. 401–417. https://doi.org/10.1007/978-3-319-45348-4_23

[48] Irene Teinemaa, Marlon Dumas, Marcello La Rosa, and Fabrizio Maria Maggi. 2017. Outcome-Oriented Predictive Process Monitoring: Review and Benchmark. *CoRR* abs/1707.06766 (2017). arXiv:1707.06766 http://arxiv.org/abs/1707.06766

[49] Ruben Urraca-Valle, Javier Antoñanzas, Fernando Antoñanzas-Torres, and Francisco J. Martínez de Pisón Ascacibar. 2016. Estimation of Daily Global Horizontal Irradiation Using Extreme Gradient Boosting Machines *(International Joint Conference SOCO'16-CISIS'16-ICEUTE'16 - San Sebastián, Spain, October 19th-21st, 2016, Proceedings)*. 105–113. https://doi.org/10.1007/978-3-319-47364-2_11

[50] Wil M. P. van der Aalst. 2016. *Process Mining - Data Science in Action, Second Edition*. Springer. https://doi.org/10.1007/978-3-662-49851-4

[51] Wil M. P. van der Aalst, Vladimir A. Rubin, H. M. W. Verbeek, Boudewijn F. van Dongen, Ekkart Kindler, and Christian W. Günther. 2010. Process mining: a two-step approach to balance between underfitting and overfitting. *Software and System Modeling* 9, 1 (2010), 87–111. https://doi.org/10.1007/s10270-008-0106-z

[52] Wil M P van der Aalst, M Helen Schonenberg, and Minseok Song. 2011. Time prediction based on process mining. *Information Systems* 36, 2 (2011), 450–475.

[53] Sjoerd van der Spoel, Maurice van Keulen, and Chintan Amrit. 2012. Process prediction in noisy data sets: a case study in a dutch hospital *(International Symposium on Data-Driven Process Discovery and Analysis)*. Springer, 60–83.

[54] Boudewijn F van Dongen, Ronald A Crooy, and Wil M P van der Aalst. 2008. Cycle time prediction: when will this case finally be finished? *(CoopIS)*. Springer, 319–336.

[55] Ilya Verenich, Hoang Nguyen, Marcello La Rosa, and Marlon Dumas. 2017. White-box prediction of process performance indicators via flow analysis. In *Proceedings of the 2017 International Conference on Software and System Process, Paris, France, ICSSP 2017, July 5-7, 2017*. 85–94. https://doi.org/10.1145/3084100.3084110